\documentclass{article} 
\usepackage{iclr2027_conference,times}

\usepackage{amsmath,amsfonts,bm}

\def\eqref#1{equation~\ref{#1}}

\def\1{\bm{1}}

\DeclareMathAlphabet{\mathsfit}{\encodingdefault}{\sfdefault}{m}{sl}
\SetMathAlphabet{\mathsfit}{bold}{\encodingdefault}{\sfdefault}{bx}{n}

\usepackage{hyperref}
\usepackage{url}

\usepackage{graphicx}
\usepackage{booktabs}
\usepackage[table]{xcolor}
\usepackage{tcolorbox}
\usepackage{colortbl}
\usepackage{booktabs}
\definecolor{headerblue}{RGB}{30,60,120}
\definecolor{rowblue}{RGB}{220,230,255}
\definecolor{rowgreen}{RGB}{220,245,220}
\definecolor{rowred}{RGB}{255,230,230}
\definecolor{rowyellow}{RGB}{255,245,210}
\usepackage{float}
\definecolor{opensource}{RGB}{220,235,255}   
\definecolor{reasoning}{RGB}{220,245,220}    
\definecolor{closed}{RGB}{255,230,230}
\definecolor{headerpurple}{RGB}{60,40,120}
\definecolor{lightpurple}{RGB}{235,230,255}
\definecolor{lightgreen}{RGB}{225,245,225}
\definecolor{lightblue}{RGB}{225,240,255}
\usepackage{subcaption}
\definecolor{headerblue}{RGB}{30,60,130}
\definecolor{rowblue}{RGB}{230,240,255}
\definecolor{rowgray}{RGB}{245,245,245}
\definecolor{highlightgreen}{RGB}{220,245,220}

\definecolor{lightgray}{RGB}{245,245,245}
\definecolor{headblue}{RGB}{220,230,245}
\definecolor{myblue}{RGB}{30,90,160}
\definecolor{lightblue}{RGB}{230,240,255}
\definecolor{mygreen}{RGB}{20,140,90}
\definecolor{myred}{RGB}{200,60,60}
\usepackage{wrapfig}
\usepackage{algorithm}
\usepackage{algpseudocode}
\usepackage{footmisc}
\usepackage{xspace}
\usepackage[accsupp]{axessibility}  

\title{QTrack: Learning to Use Temporal Evidence for Reference-Grounded Video Tracking}

\author{
Tajamul Ashraf$^{1,3}$\thanks{Corresponding author: \texttt{tashraf4@jhu.edu}}, \;
Tavaheed Tariq$^{3}$\thanks{These authors contributed equally.}, \;
Sonia Yadav$^{3}$\footnotemark[2], \;
Abrar Ul Riyaz$^{3}$\footnotemark[2], \\
\textbf{Adab Ismail}$^{3}$, \;
\textbf{Moloud Abdar}$^{2}$, \;
\textbf{Janibul Bashir}$^{3}$ \\[0.5em]
$^{1}$Johns Hopkins University, USA \\
$^{2}$The University of Queensland, Australia \\
$^{3}$Gaash Research Lab, National Institute of Technology Srinagar, India
}
\def\mot{\texttt{MOT}\xspace}
\def\qtrack{\textbf{QTrack}\xspace}
\def\rmot{\texttt{RMOT26}\xspace}
\def\vlms{\texttt{VLMs}\xspace}

\def\tapo{\texttt{TAPO}\xspace}
\def\papo{\texttt{PAPO}\xspace}
\def\grpo{\texttt{GRPO}\xspace}
\def\idf{\texttt{IDF1}\xspace}

\def\hota{\texttt{HOTA}\xspace}

\def\motp{\texttt{MOTP}\xspace}

\iclrfinalcopy 
\begin{document}

\maketitle
\lhead{Preprint}          
\begin{abstract}
Category-level multi-object trackers estimate trajectories for all detected instances, whereas some applications require propagating only targets specified by a user. We study offline, reference-grounded video tracking: given an ordered video clip, a natural-language query, and the reference-frame boxes of the specified targets, a model predicts their subsequent trajectories. We introduce \textbf{QTrack}, a vision-language policy trained with structured trajectory rewards and Temporal Perception-Aware Policy Optimization (TAPO). TAPO contrasts the likelihood of a generated response under the observed clip and a reference-frame repetition, encouraging the output to depend on visual evidence beyond the reference frame. We also introduce \rmot, a controlled diagnostic benchmark constructed from existing tracking datasets. Under the reported protocol, QTrack obtains a motion-consistency score of 0.30, MOTP of 0.75, center-location error of 44.61 pixels, and normalized distance error of 0.39; its motion-consistency score ties the strongest comparison method. Component ablations suggest complementary effects from the motion reward and temporal contrast, while an exploratory four-clip, 50-frame evaluation examines transfer beyond the primary short-clip setting. The results support further study of temporal-evidence learning in
trajectory-generating vision-language models. Code and data are available at
\href{https://anonymous.4open.science/r/QTrack-5548}{https://anonymous.4open.science/r/QTrack-5548}.
\end{abstract}

\section{Introduction}

Video tracking estimates the locations of persistent objects across frames. Category-level multi-object tracking usually detects and associates every instance of a predefined class, while single-object and referring trackers follow a selected target~\cite{bernardin2008evaluating, kalake2021analysis, luo2021multiple, shim2025focusing}. We study a controlled vision-language setting between these formulations: the input provides the reference-frame boxes of one or more specified targets, and a natural-language query supplies descriptive context; the output is a separate trajectory for each grounded target. The scientific question is whether a trajectory-generating vision-language policy can use observations beyond the reference frame to propagate these identities accurately.


Consider \textit {``a crowded clip containing similarly dressed pedestrians."} The
reference boxes identify two requested people, while the query describes their
clothing and relative positions. Predicting their later boxes requires the model
to use changing spatial evidence, distinguish nearby distractors, and preserve
the mapping between each supplied identity and its output trajectory. A model
that copies the initial boxes or relies only on static appearance can produce
plausible but incorrect trajectories as the people move or overlap.


\begin{figure}[t]
    \centering
    \includegraphics[width=\linewidth]{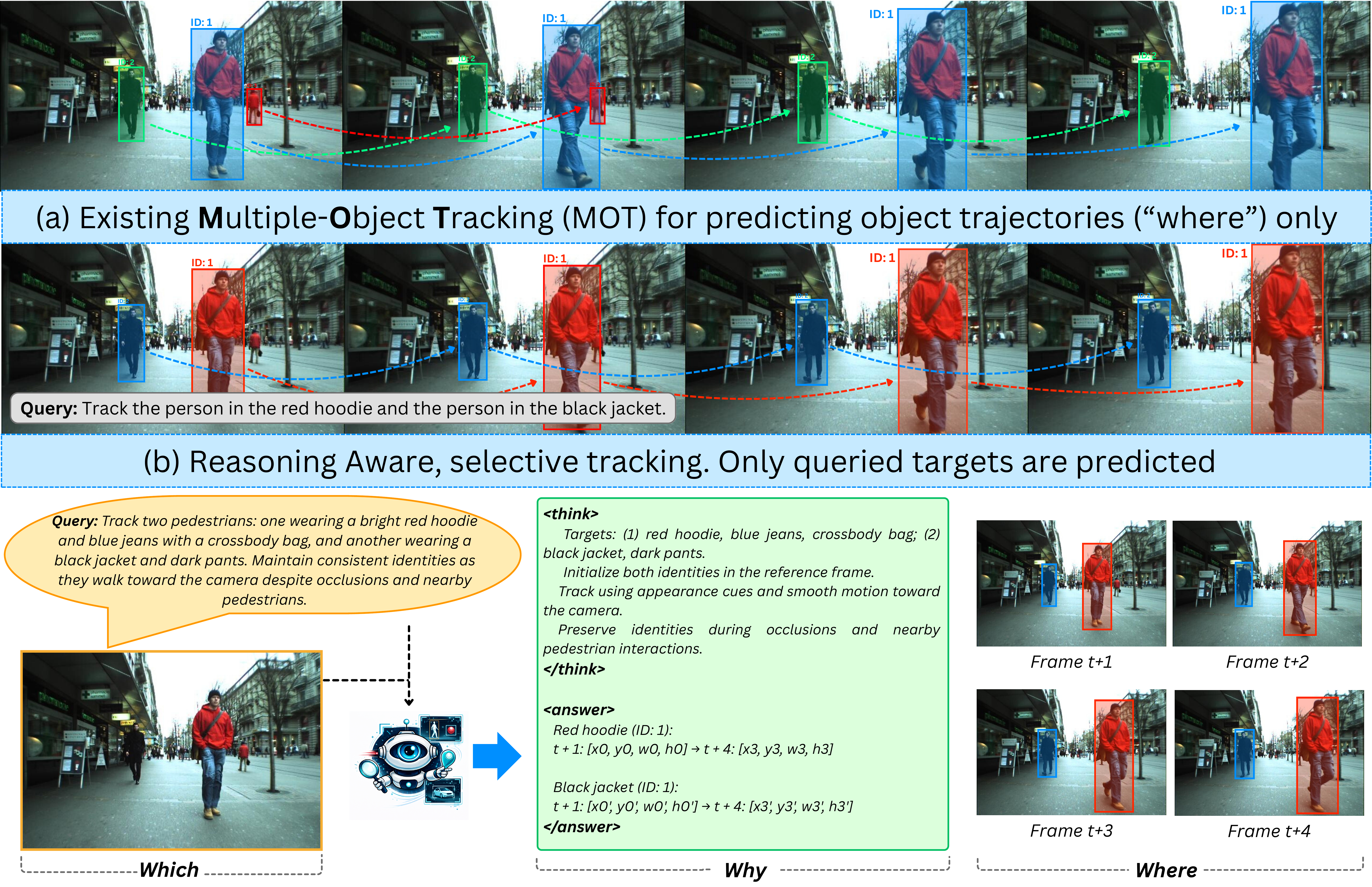}
    \caption{\textbf{Reference-grounded, query-conditioned tracking.} Category-level MOT outputs trajectories for all detected category instances. In our protocol, the input supplies reference-frame boxes for the targets specified by a query, and the model predicts one persistent trajectory per supplied target. The primary challenge is temporal propagation under distractors and visibility changes.}
    \label{fig:intro}
\end{figure}

We call this task \textit{reference-grounded, query-conditioned tracking}. Target membership is supplied through the reference boxes in the present
protocol; language can contribute attributes, relations, and motion descriptions
that help maintain the correct mapping over time as shown in Figure~\ref{fig:intro}. We therefore evaluate two questions separately: whether the policy propagates the grounded targets using temporal visual evidence, and whether the semantic content of the query improves that propagation when the boxes and video are held fixed. We do not treat this setting as language-only target discovery.


Vision-language models (\vlms) provide a natural foundation for this setting because they align visual content with free-form language and have demonstrated strong capabilities in multimodal grounding~\cite{tong2024cambrian, zhang2024llava}, visual reasoning~\cite{tan2026reason, wei2026open, al2024unibench}, and instruction following~\cite{ding2025mm, zhou2023instruction}. A coordinate-generating model may ignore later frames, copy the reference box, rematch similar objects independently in every frame, or produce well-formed outputs with spatial drift. These failure modes motivate training signals that separately measure output validity, localization, identity-preserving trajectory structure, and dependence on temporal evidence.

\textbf{Model:} We introduce \qtrack, an offline vision-language policy that generates a structured response containing a trajectory for each reference-grounded target. \qtrack is initialized from VisionReasoner-3B and optimized with Group Relative Policy Optimization (\grpo) using output-format, localization, and motion rewards. We additionally introduce Temporal Perception-Aware Policy Optimization (\tapo), a temporal adaptation of corrupted-input likelihood contrast: for the same sampled output, the model compares its likelihood under the observed clip with its likelihood when all future frames are replaced by the reference frame. The corrupted input retains the reference image and initial grounding but removes subsequent changes in position, visibility, and appearance. Because this contrast measures input sensitivity rather than correct motion by itself, we evaluate trajectory quality and temporal interventions separately.

\textbf{Data:} We release \rmot, a controlled diagnostic benchmark assembled from existing tracking datasets. Its primary examples are short, reference-grounded clips used to test localization, query use, and motion-sensitive propagation. Each instance contains a reference frame with explicit target grounding, a short video sequence, and a natural-language query specifying the tracking objective. \rmot includes single-target, multi-target, and occlusion-aware queries across diverse real-world scenarios, including crowded scenes, sports interactions, non-linear motion, and partial occlusion.

Under the current \rmot protocol, \qtrack combines the strongest reported localization results with a motion-consistency score that ties the best comparison method. Matched component ablations indicate that spatial \grpo training, the motion reward, and \tapo have different effects. A fixed-box language intervention tests whether query content contributes beyond the supplied coordinates, and a trajectory-only ablation tests the cost and benefit of generating structured reasoning. We report these results with their protocols and avoid using rationale-quality correlation as evidence of causal faithfulness. In summary, our contributions are as follows.
\begin{itemize}
    \item We define a controlled, reference-grounded tracking protocol for studying
trajectory generation by a vision-language policy and release \rmot with
predefined sequence-level splits, verified queries, and a documented evaluation
protocol.
    
    \item We introduce \qtrack, which combines structured trajectory rewards with a temporal corrupted-input contrast designed to increase dependence on observations beyond the reference frame.

    \item We separate the effects of task-specific training, motion reward, temporal contrast, language content, and structured reasoning through matched ablations, and evaluate the resulting model against general vision-language models and specialized tracking baselines.

\end{itemize}
\section{Related Work}
\paragraph{Tracking and language-guided grounding.}
Category-level multi-object tracking estimates trajectories for all instances of
a predefined class, using detection, association, or joint tracking
architectures~\citep{bewley2016simple,wojke2017simple,zhang2021fairmot,dendorfer2021motchallenge}.
Language-guided single-object tracking instead specifies a target with a natural
language expression~\citep{li2017tracking,wang2021towards}, while referring
multi-object tracking extends language guidance to multiple objects
~\citep{wu2023referring,li2025lamot}. \qtrack studies a narrower reference-grounded
setting: the target identities are supplied by initial boxes, and the model must
generate their later trajectories. This protocol isolates localization and
short-horizon propagation; it does not test language-only target discovery.

\paragraph{Reasoning-oriented tracking.}
ReasoningTrack combines supervised fine-tuning and \grpo for long-term
single-object tracking with textual reasoning~\citep{wang2025reasoningtrack}.
ReaMOT addresses implicit social or causal instructions with a modular
Thinking-LVLM--SAM2 pipeline and full-video association~\citep{chen2025reamot}.
These settings are complementary to \qtrack: ReasoningTrack emphasizes long-term
single-target tracking, ReaMOT emphasizes implicit target discovery, and \qtrack
evaluates direct coordinate generation after reference boxes identify the
requested targets. Their published scores should not be compared directly
because their inputs, temporal horizons, output protocols, and metrics differ.

\paragraph{Policy optimization for visual reasoning.}
\grpo and related reinforcement-learning objectives have been adapted to
vision-language reasoning~\citep{shao2024deepseekmath,liu2025visual,zheng2025easyr1}.
Video-R1 applies reinforcement learning to video reasoning
~\citep{feng2025videor1reinforcingvideoreasoning}, and PAPO contrasts likelihoods under original and
perceptually corrupted visual inputs~\citep{wang2026perception}. \qtrack shares
PAPO's corrupted-input likelihood-contrast principle and specializes the
corruption to temporal evidence by repeating the reference frame. The intended
distinction is therefore the temporal intervention and tracking.

\vspace{-0.1 cm}
\section{Methodology}
\vspace{-0.2 cm}


\subsection{Preliminaries and Problem Setup}
Let $\mathcal{F}=(F_0,F_1,\ldots,F_T)$ be an ordered video clip, let $q$ be a
natural-language query, and let
\begin{equation}
\mathcal{G}_0=\{(j,b_j^0)\}_{j=1}^{N_q}
\end{equation}
contain the reference-frame box $b_j^0\in\mathbb{R}^4$ for every target requested
by the query and $N_q$ be number of query-relevant targets. Thus, $\mathcal{G}_0$ supplies target membership; the task is not
language-only target discovery. The desired output is one trajectory per supplied
identity,
\begin{equation}
\mathcal{Y}^{*}=\{\tau_j^{*}\}_{j=1}^{N_q},
\qquad
\tau_j^{*}=(b_j^{*,0},b_j^{*,1},\ldots,b_j^{*,T}).
\end{equation}
QTrack generates a structured response
\begin{equation}
o=(z,\widehat{\mathcal{Y}}),
\qquad
\widehat{\mathcal{Y}}=\{\widehat{\tau}_j\}_{j=1}^{\widehat N_q},
\qquad
\pi_\theta(o\mid\mathcal{F},q,\mathcal{G}_0),
\end{equation}
where $z$ is a textual rationale and $\widehat{\tau}_j$ is a predicted trajectory.
Missing or extra outputs are penalized during spatial matching. The current
benchmark contains positive queries with one or more supplied targets and does
not evaluate an empty-target response.

\noindent\textbf{Why a vision-language policy?}
The policy can condition trajectory generation on free-form descriptions of
attributes, relations, and motion while emitting coordinates in a single
structured response. The experiments therefore test whether this formulation
improves reference-grounded propagation; they do not establish open-vocabulary
target discovery. We train \qtrack using \grpo style optimization, which compares multiple sampled outputs for the same input and updates the policy using relative rewards. For each input $x$, the old policy $\pi_{\theta_{\mathrm{old}}}$ samples a group of $G$ rollouts:
$
\{o_i\}_{i=1}^{G}
\sim
\pi_{\theta_{\mathrm{old}}}(\cdot\mid x).
$
Each rollout receives a scalar reward $R_i=R(o_i,\mathcal{Y}^{*})$ measuring output validity, spatial accuracy, and temporal consistency.

\begin{figure}[t]
    \centering
    \includegraphics[width=\linewidth]{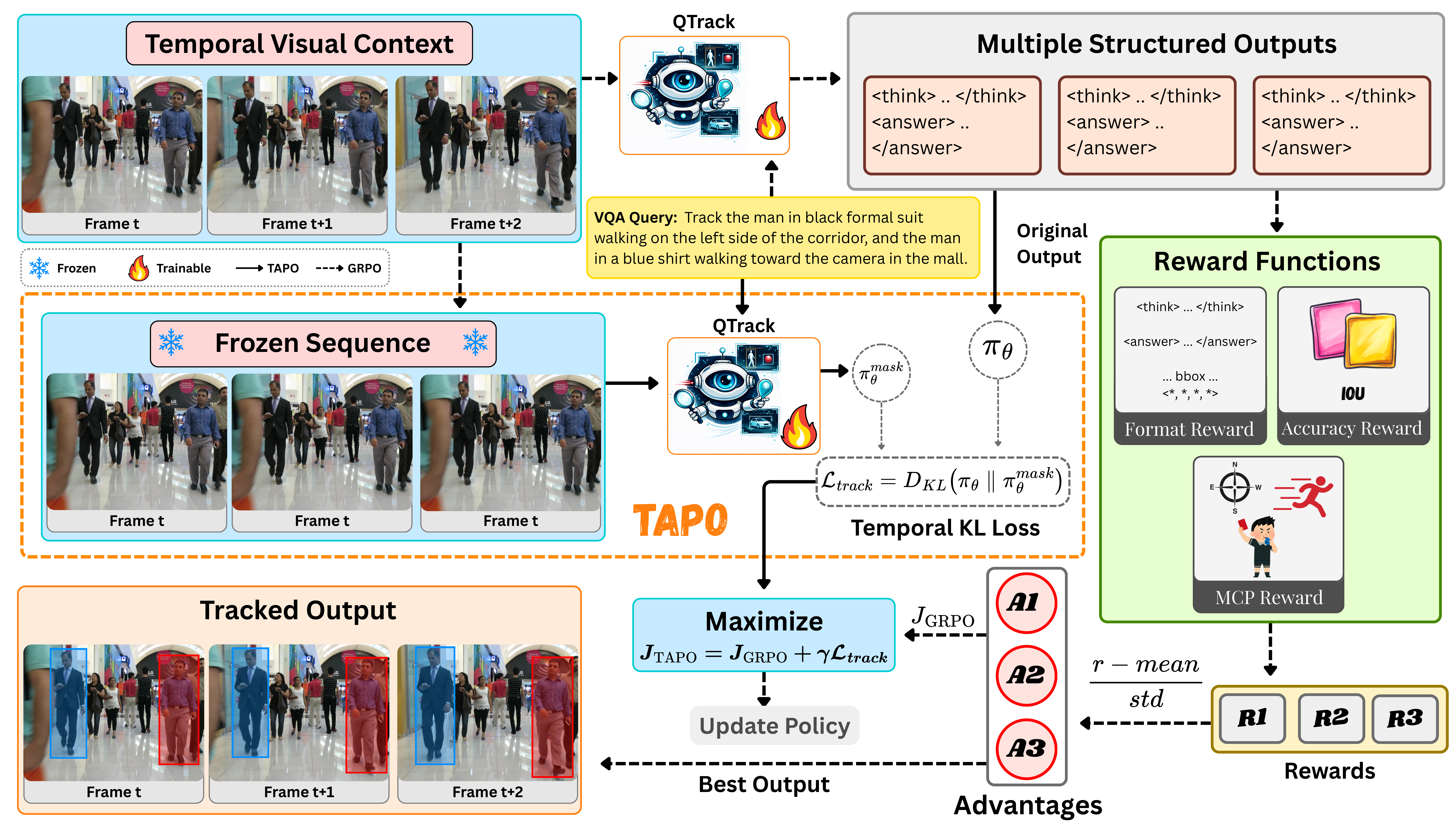}
    \caption{\textbf{Overview of the \qtrack framework.}
    Given a video sequence $\mathcal{F}=(F_0,\dots,F_T)$, reference grounding $\mathcal{G}_0$, and query $q$, \qtrack generates a structured reasoning trace and predicts bounding-box trajectories only for the query-specified target(s). The model jointly performs query-conditioned grounding, temporal association, and trajectory prediction.}
    \label{fig:main_architecture}
\end{figure}

\subsection{QTrack Framework}
\label{sec:qtrack_architecture}

Given $x=(\mathcal{F},q,\mathcal{G}_0)$, QTrack produces
\begin{equation}
(z,\widehat{\mathcal{Y}})=F_\theta(\mathcal{F},q,\mathcal{G}_0),
\qquad
\widehat{\mathcal{Y}}=\{\widehat{\tau}_j\}_{j=1}^{\widehat N_q}.
\end{equation}
The same serialization supports single-target and multi-target examples. Because
the current protocol supplies every requested identity in the reference frame,
the model is evaluated on propagation, localization, and preservation of the
identity-to-output mapping.

\subsection{Unified Reward Mechanism}
\label{sec:reward_mech}

For each rollout $o=(z,\widehat{\mathcal{Y}})$ and ground truth $\mathcal{Y}^{*}$, we use a sequence-level reward
\begin{equation}
\label{eq:total_reward}
R(o,\mathcal{Y}^{*})
=
\lambda_{\mathrm{think}}R_{\mathrm{think}}
+
\lambda_{\mathrm{fmt}}R_{\mathrm{fmt}}
+
\lambda_{\mathrm{iou}}R_{\mathrm{iou}}
+
\lambda_{\mathrm{mcp}}R_{\mathrm{MCP}} .
\end{equation}
$R_{\mathrm{think}}$ verifies that the output contains valid \texttt{<think>} and \texttt{<answer>} fields, while $R_{\mathrm{fmt}}$ checks that bounding boxes are parsable in the required JSON format. These rewards stabilize structured generation but do not directly measure tracking quality.

\noindent\textbf{Spatial reward.}
At frame $t$, predicted boxes $\widehat{B}_t$ and ground-truth boxes $B_t^{*}$ are matched using Hungarian assignment with cost $1-\mathrm{IoU}$. Let $\mathcal{M}_t$ denote the matched pairs. We define
\begin{equation}
\label{eq:iou_reward}
R_{\mathrm{iou}}
=
\frac{1}{T+1}
\sum_{t=0}^{T}
\frac{1}{\max(\widehat{N}_q,N_q)}
\sum_{(i,j)\in\mathcal{M}_t}
\mathbb{I}
\left[
\mathrm{IoU}(\widehat{b}_i^t,b_j^{*,t})>\eta
\right],
\end{equation}
with $\eta=0.5$.

\noindent\textbf{Motion reward.}
Let $c(b)$ be the center of box $b$, and define predicted and ground-truth motion vectors
$
\Delta \widehat{c}_j^t=c(\widehat{b}_j^t)-c(\widehat{b}_j^{t-1}),
\qquad
\Delta c_j^{*,t}=c(b_j^{*,t})-c(b_j^{*,t-1}).
$
For each target and timestep, we compute direction alignment, speed consistency, and anti-static penalty:
\begin{equation}
\label{eq:mcp_components}
\resizebox{\linewidth}{!}{$
A_j^t=
\frac{
\langle \Delta \widehat{c}_j^t,\Delta c_j^{*,t}\rangle
}{
\|\Delta \widehat{c}_j^t\|_2\|\Delta c_j^{*,t}\|_2+\epsilon
},
\quad
S_j^t=
\exp\left(
-\frac{
(\|\Delta \widehat{c}_j^t\|_2-\|\Delta c_j^{*,t}\|_2)^2
}{
2\sigma_v^2
}
\right),
\quad
P_j^t=
\begin{cases}
1, & \|\Delta c_j^{*,t}\|_2 \leq \epsilon_v,\\
1-\exp\left(-\frac{\|\Delta \widehat{c}_j^t\|_2^2}{2\sigma_s^2}\right), & \text{otherwise}.
\end{cases}
$}
\end{equation}
The Motion Consistency Precision (MCP) reward is
\begin{equation}
\label{eq:mcp_reward}
R_{\mathrm{MCP}}
=
\frac{1}{N_qT}
\sum_{j=1}^{N_q}
\sum_{t=1}^{T}
\max(0,A_j^t)S_j^tP_j^t .
\end{equation}
This term penalizes static boxes, incorrect motion direction, and unstable speed. The anti-static term prevents the optimizer from discovering a degenerate solution in which it predicts nearly stationary boxes for targets that are moving. During training, MCP is only one component of the reward in Eq.~\ref{eq:total_reward}; evaluation reports MCP together with MOTP, CLE, and NDE to avoid optimizing a single metric in isolation.

\subsection{Temporal Perception-Aware Policy Optimization}
\label{sec:tapo}
\begin{wrapfigure}{r}{0.50\textwidth}
\vspace{-0.8em}
\begin{minipage}{0.50\textwidth}
\footnotesize
\hrule
\vspace{0.2em}
\captionof{algorithm}{TAPO: Temporal Perception-Aware Optimization}
\label{alg:tapo_algo}
\vspace{-0.5em}
\hrule
\vspace{0.3em}

\begin{algorithmic}[1]
\State \textbf{Input:} Video $\mathcal{F}$, query $q$, grounding $\mathcal{G}_0$, policy $\pi_\theta$, weight $\gamma$
\State Sample rollout $o \sim \pi_\theta(\cdot \mid q,\mathcal{F},\mathcal{G}_0)$
\State Construct $\mathcal{F}^{\mathrm{mask}}=(F_0,F_0,\dots,F_0)$
\State Compute original log-probability
\State \hspace{0.4cm} $\log p = \log \pi_\theta(o \mid q,\mathcal{F},\mathcal{G}_0)$
\State Compute corrupted log-probability
\State \hspace{0.4cm} $\log p^{\mathrm{mask}} = \log \pi_\theta(o \mid q,\mathcal{F}^{\mathrm{mask}},\mathcal{G}_0)$
\State Estimate temporal log-ratio
\State \hspace{0.4cm} $\widehat{\mathcal{L}}_{\mathrm{track}}=\log p-\log p^{\mathrm{mask}}$
\State $J_{\mathrm{TAPO}}=J_{\mathrm{GRPO}}+\gamma\widehat{\mathcal{L}}_{\mathrm{track}}$
\State Update $\theta$ using $J_{\mathrm{TAPO}}$
\end{algorithmic}

\vspace{0.3em}
\hrule
\end{minipage}
\vspace{-1.2em}
\end{wrapfigure}
Although \grpo improves structured output quality, vision-language trackers can still over-rely on static appearance cues. In crowded scenes, this can produce trajectories that are plausible frame-by-frame but insensitive to motion, leading to drift under occlusion or similar-looking distractors. We address this with Temporal Perception-Aware Policy Optimization (\tapo), which explicitly encourages the policy to distinguish motion-bearing videos from temporally corrupted ones.

\noindent\textbf{Temporal corruption.}
Given a video sequence $\mathcal{F}=(F_0,F_1,\dots,F_T)$, we construct a motion-corrupted sequence
\begin{equation}
\mathcal{F}^{\mathrm{mask}}=(F_0,F_0,\dots,F_0),
\qquad
F_t^{\mathrm{mask}}=F_0,\ \forall t.
\end{equation}
This preserves static appearance and reference-frame evidence while removing temporal motion. A motion-aware policy should therefore assign different likelihoods to trajectory rollouts under $\mathcal{F}$ and $\mathcal{F}^{\mathrm{mask}}$.

\noindent\textbf{Temporal log-ratio estimator.}
We define the temporal sensitivity objective as a forward KL divergence from the policy conditioned on the original video to the policy conditioned on the corrupted video:
\begin{equation}
\label{eq:l_track}
\mathcal{L}_{\mathrm{track}}
=
D_{\mathrm{KL}}
\left(
\pi_{\theta}(\cdot \mid q,\mathcal{F},\mathcal{G}_0)
\;\middle\|\;
\pi_{\theta}(\cdot \mid q,\mathcal{F}^{\mathrm{mask}},\mathcal{G}_0)
\right).
\end{equation}
In practice, we estimate this term using sampled rollouts. For a rollout $o$ generated from the original-video policy, the single-sample Monte Carlo estimator is
\begin{equation}
\label{eq:l_track_mc}
\widehat{\mathcal{L}}_{\mathrm{track}}
=
\log \pi_{\theta}(o \mid q,\mathcal{F},\mathcal{G}_0)
-
\log \pi_{\theta}(o \mid q,\mathcal{F}^{\mathrm{mask}},\mathcal{G}_0).
\end{equation}
Thus, Algorithm~\ref{alg:tapo_algo} implements a sampled log-ratio estimate of the KL in Eq.~\ref{eq:l_track}, rather than a separate full-distribution KL computation. Maximizing this term discourages motion-invariant behavior: a policy should not assign the same likelihood to a trajectory when all future frames have been frozen.

\noindent\textbf{Overall objective.}
Since \grpo is written as a maximization objective, the final objective is
\begin{equation}
\label{eq:tapo_objective}
J_{\mathrm{TAPO}}(\theta)
=
J_{\mathrm{GRPO}}(\theta)
+
\gamma
\mathbb{E}_{x}
\left[
\widehat{\mathcal{L}}_{\mathrm{track}}
\right],
\end{equation}
where $\gamma$ controls the strength of temporal regularization. In loss-minimization form, this is equivalent to
$
\mathcal{L}_{\mathrm{TAPO}}
=
-
J_{\mathrm{GRPO}}
-
\gamma
\widehat{\mathcal{L}}_{\mathrm{track}}.
$

\noindent\textbf{Relation to perception-aware regularization.}
Unlike generic corrupted-input consistency and \papo~\cite{wang2026perception}, \tapo removes temporal evidence while preserving appearance. Its log-ratio regularizer is applied to trajectory rollouts within \grpo, explicitly penalizing motion-invariant policies for identity-preserving video tracking.
\section{RMOT26 Benchmark}
\label{sec:rmot26}
\vspace{-0.2 cm}

\begin{figure}[t]
    \centering
    \includegraphics[width=\linewidth]{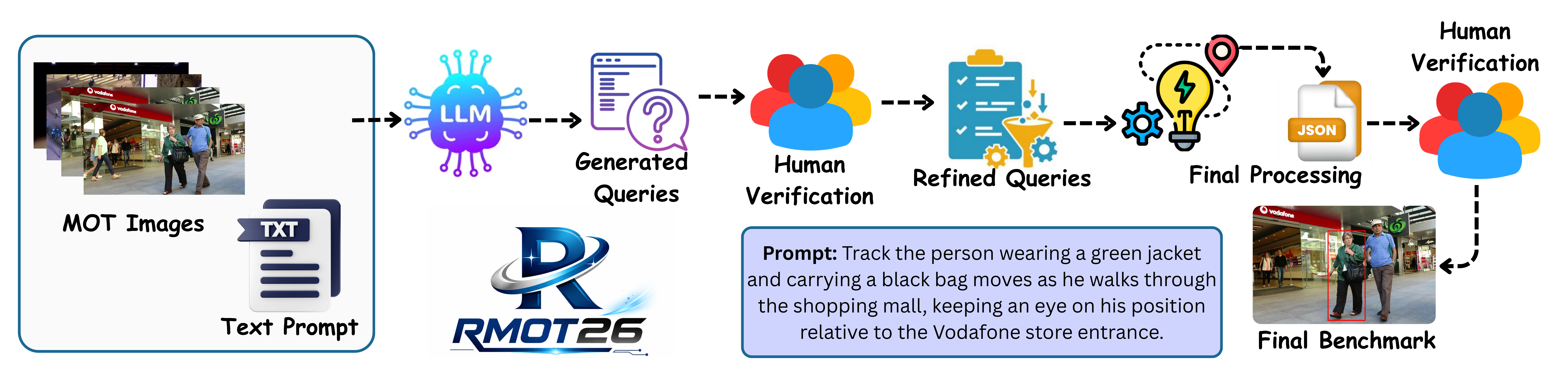}
    \caption{\textbf{RMOT26 construction pipeline.}
    We sample short video windows from existing tracking datasets, select query-relevant targets using reference-frame annotations, generate grounded natural-language queries, verify target identities and trajectories, and convert each instance into a structured query-conditioned tracking sample.}
    \label{fig:dataset_pipeline}
\end{figure}

\noindent\textbf{Dataset design.}
\rmot is a diagnostic benchmark for query-conditioned spatiotemporal grounding in videos. Each instance is defined as
$
\mathcal{I}
=
(\mathcal{F}, \mathcal{G}_0, q, \mathcal{Y}^{*}),
$
where $\mathcal{F}=(F_0,F_1,\dots,F_T)$ is a short video clip, $\mathcal{G}_0$ denotes explicit reference-frame grounding, $q$ is a natural-language query, and $\mathcal{Y}^{*}=\{\tau_j^{*}\}_{j=1}^{N_q}$ is the set of ground-truth trajectories for the query-relevant target(s). Each trajectory is
$
\tau_j^{*}
=
(b_j^{*,0}, b_j^{*,1},\dots,b_j^{*,T}),
$
where $b_j^{*,t}$ is the bounding box of target $j$ in frame $F_t$.
We curate clips from seven tracking datasets: DanceTrack~\cite{sun2022dancetrack}, MOT16/17~\cite{milan2016mot16, leal2015motchallenge}, MOT20~\cite{dendorfer2020mot20}, SportsMOT~\cite{cui2023sportsmot}, DAMUNT~\cite{abeysinghe2023tracking} and PETS~\cite{ferryman2009pets2009}. These sources cover diverse motion regimes, including non-linear pedestrian motion, fast sports interactions, crowded surveillance scenes, heavy overlaps, and static-camera settings. This diversity allows \rmot to test whether models can move beyond frame-level grounding toward motion-aware query-conditioned tracking. Benchmarks such as LaSOT~\citep{fan2019lasot}, TNL2K~\citep{wang2021towards}, and TNLLT~\citep{wang2025reasoningtrack} focus on single-target language-guided tracking, whereas RMOT26 supports single- or multi-target queries with identity-consistent trajectories, penalizes missing and extra trajectories, jointly evaluates spatial localization and motion consistency, and uses explicit reference grounding to isolate temporal propagation, making the settings complementary: LaSOT/TNL2K-style benchmarks emphasize long-horizon single-object tracking, while RMOT26 targets controlled, query-conditioned multi-target trajectory generation.
The claims supported by RMOT26 are: (1) propagation of one or more
targets supplied by reference-frame boxes; (2) localization over the benchmark's
controlled horizon; (3) short-horizon preservation of the supplied
identity-to-output mapping under crowding, interaction, and partial occlusion;
and (4) sensitivity to temporal evidence under the reported intervention. RMOT26
does not evaluate language-only target discovery, online tracking, or long-term
re-identification.


\noindent\textbf{Query categories.}
Each query is grounded in the reference frame and maps to one or more target trajectories. We define three evaluation categories:
\textit{Single-object queries} specify one target in the reference frame and require predicting its trajectory across future frames.
\textit{Multi-object queries} specify multiple targets and require producing separate identity-consistent trajectories for each queried object.
\textit{Occlusion-aware queries} involve targets that are partially occluded or visually ambiguous, requiring identity preservation under visibility changes.
Because every requested identity is spatially anchored by a reference-frame box,
errors primarily reflect future localization, temporal propagation, output
parsing, or preservation of the identity-to-output mapping.

\noindent\textbf{Dataset scale.}
The test split contains 900 single-object and 100 multi-object instances. Although the multi-object subset is smaller, it is substantially more annotation-intensive: each instance requires separate reference grounding, target-specific language descriptions, and consistent identity labels for every queried object. Each multi-object sequence contains 2--5 tracked targets, yielding more than 300 annotated trajectory-query pairs in the multi-target subset.

\noindent\textbf{Annotation protocol.}
Trajectory supervision is derived from verified tracking annotations. For each selected target, we extract its bounding boxes across the reference and future frames to form an ordered trajectory. Single-object samples are represented as one trajectory array, while multi-object samples use a structured identity-to-trajectory mapping:
$
\mathcal{Y}^{*}
=
\{(id_j,\tau_j^{*})\}_{j=1}^{N_q}.
$
Each sample contains: (1) a natural-language description of the target(s), (2) explicit reference-frame bounding boxes, and (3) a query prompt specifying spatial, temporal, or relational constraints.
To improve linguistic diversity, query templates are generated with Qwen-2.5-VL-Instruct~\cite{qwen2.5-VL}, conditioned on scene context, object attributes, and motion cues. All generated queries are verified against the reference grounding and original identity annotations, ensuring that the benchmark remains visually grounded and does not rely on external commonsense assumptions. Additional details are provided in Appendix~\S\ref{appendix:dataset_statistics}.


\section{Experiments}
\label{sec:experiments}
\vspace{-0.2 cm}

We evaluate \qtrack along four axes: spatial localization, short-horizon motion
consistency, contributions of the reported training components, and
rationale-trajectory consistency. All experiments use RMOT26 unless otherwise
stated.

\begin{table}[tb]
\centering
\caption{\textbf{Main results on the RMOT26 benchmark.} Higher is better for MCP and MOTP, while lower is better for CLE and NDE.}
\label{tab:rmot26_main_results}

\resizebox{\textwidth}{!}{
\begin{tabular}{@{}l l l l l l@{}}
\toprule
\textbf{Model Name} & \textbf{Params} & \textbf{MCP$\uparrow$} & \textbf{MOTP$\uparrow$} & \textbf{CLE (px)$\downarrow$} & \textbf{NDE$\downarrow$} \\
\midrule

\rowcolor{opensource}
\multicolumn{6}{l}{\textbf{Open-Source Models}} \\
\midrule
Qwen3-VL-Instruct~\cite{yang2025qwen3} & 8B  & 0.25 & 0.64 & 96.0 & 0.97 \\
Gemma 3~\cite{team2025gemma} & 27B & 0.24 & 0.56 & \underline{58.4} & 0.88 \\
Gemma 3~\cite{team2025gemma} & 12B & 0.18 & \underline{0.73} & 172.9 & 0.95 \\
Llama 3.2 Vision-Instruct~\cite{grattafiori2024llama} & 11B & 0.19 & 0.15 & 552.1 & 2.67 \\
DeepSeek~\cite{wu2024deepseek} & 16B & 0.11 & 0.27 & 989.13 & 4.7 \\
Mistral-3-Instruct~\cite{liu2026ministral} & 8B & 0.15 & 0.54 & 225.3 & 0.98 \\
Molmo 2~\cite{clark2026molmo2} & 8B & \underline{0.28} & 0.54 & 271.87 & 1.05 \\
Ovis 2.5~\cite{lu2025ovis2} & 9B & 0.10 & 0.08 & 986.7 & 4.74 \\
\midrule

\rowcolor{reasoning}
\multicolumn{6}{l}{\textbf{Reasoning Models}} \\
\midrule
VisionReasoner~\cite{liu2025visionreasoner} & 7B & 0.23 & 0.24 & 428.9 & 2.24 \\
VisionReasoner~\cite{liu2025visionreasoner} & 3B & 0.21 & 0.44 & 416.58 & 2.32 \\
Migician~\cite{li2025migician} & 7B & 0.25 & 0.22 & 658.39 & 3.28 \\
InternVL~\cite{chen2024internvl} & 8B & 0.21 & 0.66 & 117.44 & 0.64 \\
\midrule

\rowcolor{closed}
\multicolumn{6}{l}{\textbf{Closed-Source Models}} \\
\midrule
gpt-4o-mini~\cite{achiam2023gpt} & - & 0.20 & 0.57 & 130.48 & 0.67 \\
gpt-5.2~\cite{singh2025openai} & - & 0.25 & 0.61 & 94.2 & \underline{0.55} \\
\midrule

\rowcolor{reasoning}
\multicolumn{6}{l}{\textbf{Other Models}} \\
\midrule
TempRMOT~\cite{zhang2024bootstrapping} & - & \textbf{0.30} & 0.19 & 510 & 3.02 \\
DKGTrack~\cite{li2025language} & - & 0.28 & 0.16 & 580 & 3.35 \\
\midrule

\textbf{QTrack (Ours)} & 3B & \textbf{0.30} & \textbf{0.75} & \textbf{44.61} & \textbf{0.39} \\
\bottomrule
\end{tabular}
}
\end{table}
\noindent\textbf{Evaluation metrics.}
All stored boxes use the $[x,y,w,h]$ convention. Center-based metrics first
convert a box to $c(b)=(x+w/2,y+h/2)$. We report MOTP, center-location error
(CLE), normalized displacement error (NDE), and Motion Consistency Precision
(MCP).


\noindent\textbf{Implementation details.}
All experiments are run on NVIDIA H100 GPUs. \qtrack is initialized from the 3B VisionReasoner backbone~\cite{liu2025visionreasoner} and trained with \grpo using $G=8$ rollouts per input.
Each output is serialized as a structured response containing a reasoning field and a JSON trajectory field:
\texttt{<think>...</think><answer>[id, frame, bbox\_2d]</answer>}.
Invalid or unparsable outputs receive zero format reward and zero tracking reward. For multi-target queries, predictions and ground truth are matched using Hungarian assignment; missing and extra trajectories are penalized through the $\max(\widehat{N}_q,N_q)$ normalization in Eq.~\ref{eq:iou_reward}. Closed-source and open-source baselines receive identical prompts and one decoded output per sample. Additional prompts, hyperparameters, and parsing rules are provided in Appendix~\S\ref{appendix:implementation_details}.

\subsection{Main Results on RMOT26}
\label{sec:main_results}
\vspace{-0.1 cm}

Table~\ref{tab:rmot26_main_results} compares \qtrack with open-source, closed-source, and reasoning-oriented vision-language models. \qtrack attains an MCP score of $0.30$, tying TempRMOT for the highest
reported MCP. Among the evaluated
general-purpose vision-language models, the strongest competing MCP is
$0.28$. \qtrack additionally obtains the highest MOTP and lower
center-localization and normalized-distance errors in this evaluation. These gains show that temporal consistency is not obtained by scale alone: for example, Gemma-3 27B reaches MCP $0.24$, whereas Qwen-3-VL 8B reaches MCP $0.25$, indicating that motion-aware supervision and temporal regularization are critical. For completeness, we also evaluate \qtrack with standard tracking metrics such as \idf and \hota on query-conditioned subsets of MOT17~\cite{milan2016mot16} and DanceTrack~\cite{sun2022dancetrack} in Appendix Table~\ref{tab:mot_17_dancetrack}, where it remains competitive, suggesting that selective language-guided tracking preserves trajectory quality. Compared with long-horizon language-guided trackers such as ReasoningTrack~\cite{wang2025reasoningtrack}, \rmot targets a different regime: short-horizon query-conditioned tracking with explicit target grounding and motion consistency. ReaMOT~\cite{chen2025reamot} is also complementary, using a training-free Thinking-LVLM+SAM2 pipeline for implicit full-video reasoning, whereas \qtrack is an end-to-end RL-optimized model trained with structured rewards for explicit reference-grounded tracking.

\subsection{Ablation Studies}
\label{sec:ablation}

\begin{figure}[tb]
\centering

\begin{subfigure}[t]{0.55\textwidth}
\vspace{- 2.9 cm}
\centering
\small
\rowcolors{2}{lightpurple}{white}
\begin{tabular}{lccc}
\toprule
\rowcolor{headerpurple}
\color{white}\textbf{Model (3B)} & 
\color{white}\textbf{MCP$\uparrow$} & 
\color{white}\textbf{MOTP$\uparrow$} & 
\color{white}\textbf{NDE$\downarrow$} \\
\midrule
VisionReasoner & 0.21 & 0.44 & 2.32 \\
GRPO           & 0.22 & 0.65 & 0.96 \\
MCP-Reward     & \underline{0.25} & 0.61 & 1.06 \\
TAPO           & 0.24 & \underline{0.72} & \underline{0.82} \\
\midrule
\rowcolor{lightgreen}
\textbf{QTrack (Ours)} & 
\textbf{0.30} & 
\textbf{0.75} & 
\textbf{0.39} \\
\bottomrule
\end{tabular}
\caption{Component-wise}
\end{subfigure}
\hfill
\begin{subfigure}[t]{0.4\textwidth}
\centering
\includegraphics[width=0.8\linewidth]{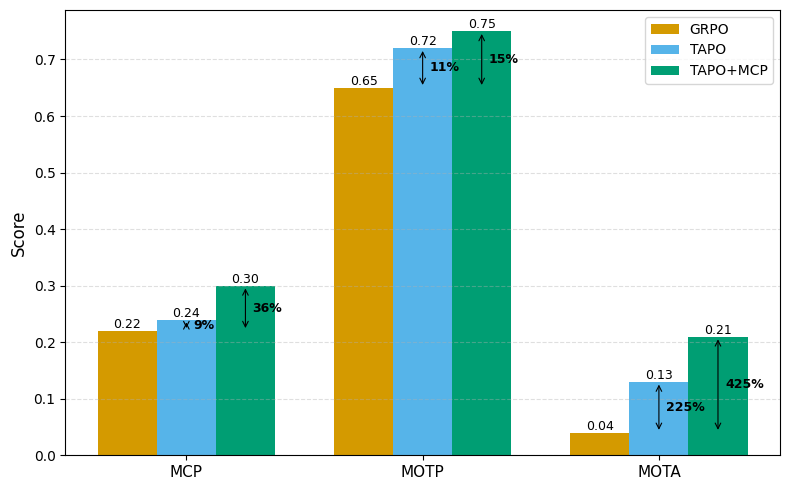}
\caption{GRPO vs TAPO comparison.}
\end{subfigure}

\caption{\textbf{Ablation analysis.}
GRPO, MCP, and \tapo provide complementary gains, with the full combination performing best.}
\label{fig:component_analysis_tapo}
\end{figure}

\noindent\textbf{Component ablation.}
Fig.~\ref{fig:component_analysis_tapo}(a) shows that the components have complementary effects rather than monotonic gains on every metric. GRPO improves localization substantially, increasing MOTP from $0.44$ to $0.65$, but MCP remains low at $0.22$, indicating limited motion consistency. Adding the MCP reward increases MCP from $0.22$ to $0.25$, while MOTP
decreases from $0.65$ to $0.61$ and NDE increases from $0.96$ to $1.06$.
This indicates that motion-oriented reward shaping introduces a trade-off
with frame-level localization accuracy when used in isolation.

\noindent\textbf{Longer-sequence evaluation.}
We additionally evaluated \qtrack on four 50-frame DanceTrack sequences (Table~\ref{tab:longer-seq}) featuring crowding, similar-looking pedestrians, and occlusion. \qtrack reduces identity switches from 29 to 21 versus VisionReasoner and achieves the best HOTA, MOTA, MOTP, IDF1, and MCP. All frames were provided simultaneously, following the clip-level protocol in Table~\ref{tab:rmot26_main_results}, so the evaluation is offline/non-causal. As the sequences were not curated for prolonged disappearance and re-entry, we view these results as evidence of longer-horizon robustness rather than a complete evaluation of online re-identification.
\begin{table*}[h]
\centering
\caption{Comparison of QTrack on longer sequences.}
\label{tab:longer-seq}

\small

\rowcolors{2}{lightgray}{white}

\begin{tabular}{lcccccc}
\toprule
\rowcolor{headblue}
\textbf{Model} & \textbf{HOTA$\uparrow$} & \textbf{MOTA$\uparrow$} & \textbf{MOTP$\uparrow$} & \textbf{IDF1$\uparrow$} & \textbf{IDSW$\downarrow$} & \textbf{MCP$\uparrow$} \\
\midrule

Qwen2.5-VL-7B & 0.281 & 0.166 & 0.708 & 0.331 & \textbf{18} & 0.334 \\
VisionReasoner-7B & 0.272 & 0.168 & 0.722 & 0.318 & 29 & 0.337 \\
QTrack-3B & \textbf{0.304} & \textbf{0.190} & \textbf{0.751} & \textbf{0.352} & \underline{21} & \textbf{0.371} \\

\bottomrule
\end{tabular}

\end{table*}

\noindent\textbf{GRPO vs. TAPO.}
Fig.~\ref{fig:component_analysis_tapo}(b) isolates the effect of \tapo. GRPO encourages higher-reward structured outputs, but the resulting trajectories can remain motion-invariant, especially in crowded scenes with similar-looking objects. \tapo addresses this failure mode by comparing model behavior under original and temporally corrupted clips, encouraging predictions that depend on true motion cues. This yields smoother trajectories, lower drift, and higher MCP.

\noindent\textbf{Comparison with traditional MOT methods.}
We conduct a restricted comparison against traditional \mot methods on query-specific clip segments from MOT17~\cite{milan2016mot16} and DanceTrack~\cite{sun2022dancetrack}. Results are reported in Appendix Table~\ref{tab:mot_comparison}, with protocol details in Appendix~\S\ref{appendix:mot_eval_protocol}. This comparison is not intended as a full-video benchmark race: all methods are evaluated on the same short query-specific clips, while \qtrack is constrained to output only query-specified targets. Under this controlled protocol, \qtrack achieves competitive or superior \hota and \motp, suggesting that selective language-conditioned tracking does not sacrifice trajectory quality. Full-video evaluation requires a different inference protocol and is left for future work.



\begin{figure}[t]
\centering
\includegraphics[width=\linewidth]{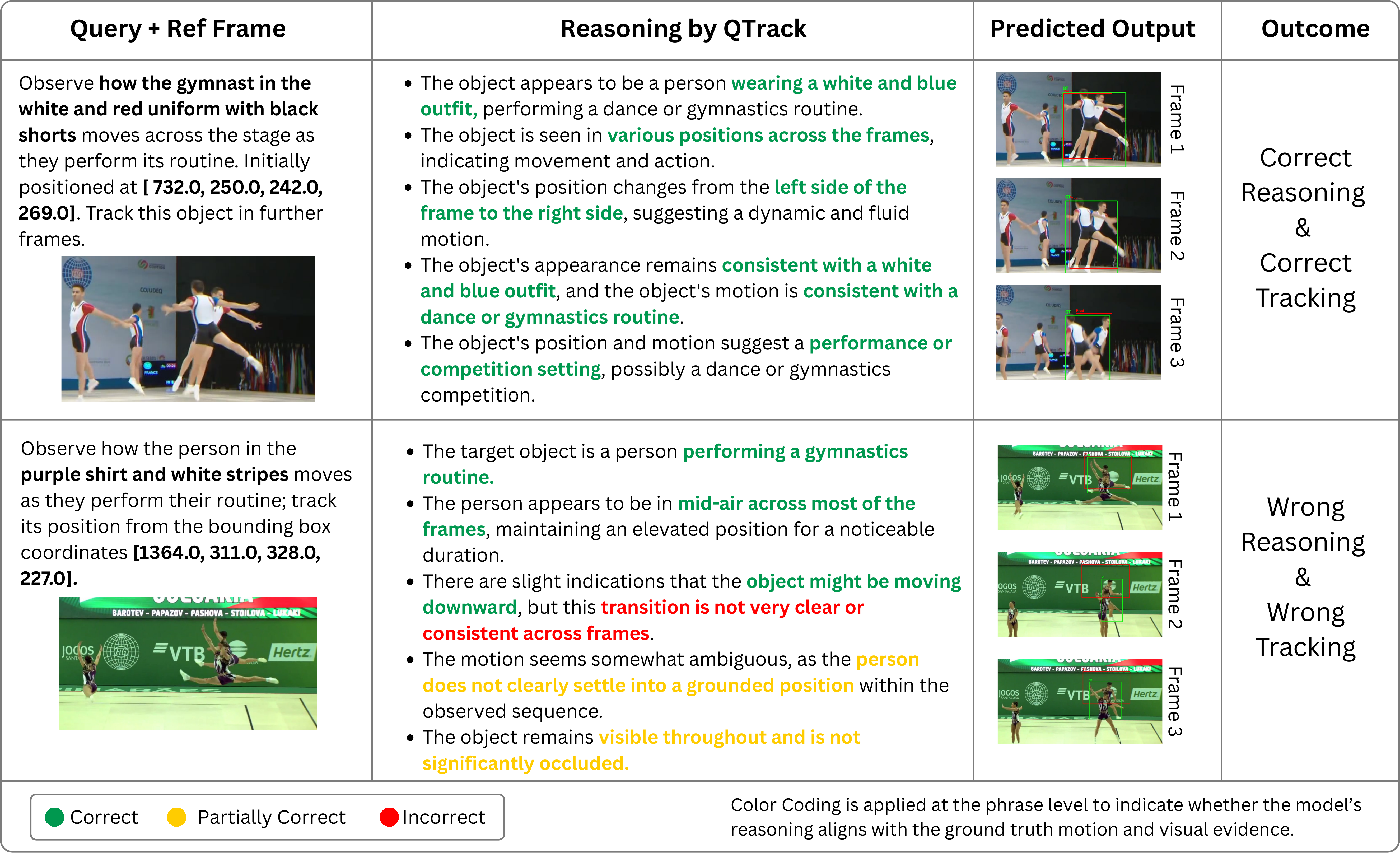}
\caption{\textbf{Reasoning trace analysis.}
Phrase-level annotations show whether the generated reasoning is consistent with visual evidence and the predicted trajectory.}
\label{fig:reasoning_qual}
\end{figure}

\noindent\textbf{Reasoning Quality Analysis.}
We score $200$ stratified cases to assess whether \qtrack's generated reasoning aligns with its trajectories. Each trace is rated on object identification, temporal reasoning, causal justification, and prediction alignment, producing a normalized Reasoning Quality Score (RQS). RQS is positively but modestly correlated with IoU ($r=+0.158$, $p=0.025$), suggesting that better reasoning traces are associated with better tracking but do not by themselves establish causality. Faithfulness decreases from $0.661$ in easy cases to $0.457$ in failure cases, indicating that low-quality predictions more often contain unsupported motion claims. The dominant failure mode is weak causal justification ($49\%$), followed by directional hallucination and motion failure ($18\%$ each). Full scoring protocol and qualitative examples are provided in Appendix~\S\ref{appendix:reasoning}.


\section{Conclusion}

We presented \qtrack, a vision--language framework for
reference-grounded, query-conditioned video tracking. We introduced
\rmot as a diagnostic benchmark for evaluating short-horizon
target propagation, localization, identity preservation, and motion
consistency. We also proposed Temporal Perception-Aware Policy
Optimization (\tapo), which uses temporally corrupted inputs and
a sampled likelihood-ratio objective to encourage predictions that depend
on evidence beyond the reference frame.
Under the reported protocol, \qtrack attains an MCP score of
$0.30$, and obtains the highest MOTP
and the lowest CLE and NDE in this evaluation. The component ablations
indicate that the motion reward and temporal contrast provide
complementary benefits, although the motion reward alone introduces a
trade-off between motion consistency and frame-level localization.
These results support further investigation of temporal-evidence
learning in trajectory-generating vision--language models.

\paragraph{Limitations.}
The primary \rmot evaluation uses short clips and therefore
does not measure long-term re-identification, re-entry after prolonged
absence, or full-video identity drift. Future work should consider longer
videos, source-video-disjoint splits, broader object categories, and
multilingual queries.

\section*{Ethics Statement}

RMOT26 is constructed from seven existing multi-object tracking benchmarks that were released for academic research. We do not collect new video recordings or introduce new personally identifying annotations. Nevertheless, some source videos depict people in public or semi-public environments, and tracking technology can create privacy risks or be misused for surveillance, profiling, or monitoring without consent. QTrack and RMOT26 are intended solely for research on reference-grounded visual reasoning and short-horizon object tracking; they are not designed or validated for identifying individuals, inferring sensitive attributes, or making high-stakes decisions. The benchmark may also inherit geographic, demographic, environmental, and behavioral biases from its source datasets, while generated language descriptions may contain appearance-related biases despite human review. These limitations should be considered when interpreting results or transferring models to new domains. Redistribution of source material will follow the applicable licenses and terms of the original datasets, and valid removal requests from the original dataset maintainers will be honored.

\section*{Reproducibility Statement}

We provide an anonymized repository containing the QTrack implementation, RMOT26 benchmark resources, evaluation code, and instructions for reproducing the reported experiments. The paper specifies the model initialization, GRPO and TAPO objectives, reward functions, training hyperparameters, hardware configuration, prompt and output formats, evaluation metrics, and baseline protocols. RMOT26 includes a data card, source-dataset composition, sequence-level train--test splits, and dataset statistics. The appendix additionally provides metric definitions, implementation details, prompt templates, traditional-tracker evaluation procedures, qualitative results, and failure analyses. The reported experiments were conducted on a single NVIDIA H100 GPU with 100\,GB of memory. Because training uses the reference frame and one future frame whereas evaluation uses sequences of five to six frames, we explicitly document this train--test sequence-length mismatch. The code and data are available at \url{https://anonymous.4open.science/r/QTrack-5548}.

\section*{AI Use Statement}

Generative AI tools were used in both the research workflow and the preparation of this manuscript. As part of the RMOT26 construction pipeline, Qwen2.5-VL-Instruct was used to generate initial natural-language queries, reasoning steps, and candidate outputs from the underlying tracking examples. These generated materials were subsequently reviewed, corrected, and refined by human annotators before inclusion in the benchmark. Large language models were also used to assist with language editing, organization, and presentation of the manuscript. AI-generated suggestions were not accepted without author review. The authors verified the dataset annotations, experimental procedures, numerical results, citations, claims, and conclusions and take full responsibility for the final submission.

\bibliography{iclr2027_conference}
\bibliographystyle{iclr2027_conference}

\appendix
\newpage
\appendix
\section*{Appendix}

\vspace{1em}

\begin{table}[h]
\centering
\renewcommand{\arraystretch}{1.35}

\resizebox{\linewidth}{!}{

\rowcolors{2}{gray!8}{white}

\begin{tabular}{@{}p{4cm} p{9.5cm}@{}}

\rowcolor{blue!20}
\textbf{Appendix Section} & \textbf{Description} \\

\toprule

\texttt{Section~\ref{datacard}} & \textbf{RMOT26 Data Card} \\

\texttt{Section~\ref{appendix:dataset_statistics}} & \textbf{RMOT26 Dataset Statistics} \\


\texttt{Section~\ref{appendix:implementation_details}} & \textbf{Metrics and Implementation Details} \\

\texttt{Section~\ref{appendix:mot_eval_protocol}} & \textbf{Evaluation Protocol for Traditional MOT Comparison} \\

\texttt{Section~\ref{appendix:finetunned_models}} & \textbf{Comparison with Fine-Tuned Models} \\

\texttt{Section~\ref{appendix:qualatative}} & \textbf{Additional Qualitative Results} \\

\texttt{Section~\ref{appendix:fail_cases}} & \textbf{Failure Cases} \\

\texttt{Section~\ref{appendix:prompt}} & \textbf{Prompt Templates for Baseline Models} \\

\texttt{Section~\ref{appendix:reasoning}} & \textbf{Reasoning Quality Evaluation Details} \\

\bottomrule

\end{tabular}
}

\label{tab:appendix_contents}

\end{table}

\section{Datacard for \rmot}
\label{datacard}

\subsection{Motivation}

\begin{itemize}

\item \textbf{For what purpose was the dataset created?}

The \rmot dataset is designed to evaluate \textit{query-driven multi-object tracking}, where models track only objects specified by a natural language query rather than all objects in a scene. Each task requires identifying the correct object(s) in a reference frame and maintaining consistent trajectories across subsequent frames. The dataset evaluates language-conditioned spatiotemporal reasoning, including query understanding, spatial grounding, temporal identity consistency, and robustness under challenging conditions such as occlusion or object interactions.

\item \textbf{Who created the dataset?}

The dataset was created by the authors of this paper.

\item \textbf{Who funded the creation of the dataset?}

Funding information will be disclosed after the anonymous review process.

\end{itemize}

\subsection{Composition}

\begin{itemize}

\item \textbf{What do the individual instances represent?}

Each instance represents a query-conditioned tracking task consisting of a reference frame with annotated bounding boxes, a sequence of subsequent frames, a natural language query describing the target object(s), and ground-truth trajectories.

\item \textbf{How many instances are there in the benchmark?}

The benchmark dataset contains 1000 instances: 900 single-object tracking tasks and 100 multi-object tracking tasks. Among the single-object tasks, 20 queries specifically focus on occlusion scenarios.

\item \textbf{Is the dataset a complete set or a sample?}

The dataset is a curated collection of tracking scenarios rather than an exhaustive set.

\item \textbf{What data does each instance contain?}

Each instance includes a reference frame, subsequent video frames, a natural language query, and ground-truth bounding box trajectories with object IDs.

\item \textbf{Is there a label associated with each instance?}

Yes. Ground-truth bounding boxes and object IDs define the target trajectories.

\item \textbf{Is any information missing from instances?}

No.

\item \textbf{Are there recommended data splits?}

Yes. The dataset provides predefined training and test splits described in Section~\ref{appendix:dataset_statistics}.

\item \textbf{Are there errors or noise in the dataset?}

The dataset is generated using a semi-automated pipeline and verified by humans. Minor annotation noise may occur due to human error.

\item \textbf{Does the dataset rely on external resources?}

Yes. The video sequences are derived from existing multi-object tracking datasets.

\item \textbf{Does the dataset contain confidential data?}

No.

\item \textbf{Does the dataset contain potentially offensive content?}

No.

\end{itemize}

\subsection{Collection Process}

\begin{itemize}

\item \textbf{How was the data acquired?}

Language queries and reasoning steps were generated using Qwen2.5-VL-Instruct~\cite{qwen2.5-VL} and subsequently reviewed and refined by human annotators. The visual data was collected from publicly available datasets.

\item \textbf{What methods were used to generate and curate the data?}

A semi-automated pipeline was used. Queries, reasoning steps, and candidate outputs were first generated by Qwen2.5-VL-Instruct and then manually verified for correctness and clarity.

\item \textbf{Who was involved in the data collection process?}

Researchers and student annotators.

\item \textbf{Over what timeframe was the data collected?}

The dataset was created between 2025 and 2026.

\item \textbf{Were ethical considerations taken into account?}

Yes. All visual data comes from datasets released for academic research. Removal requests from original dataset authors will be honored.

\end{itemize}

\subsection{Preprocessing/Cleaning/Labeling}

\begin{itemize}

\item \textbf{Was preprocessing or annotation performed?}

Yes. Automatically generated queries and reasoning outputs were manually reviewed and corrected before inclusion.

\item \textbf{Was the raw data saved?}

No separate raw data is stored. The dataset contains curated subsets of images and videos from existing datasets along with generated annotations.

\end{itemize}

\subsection{Uses}

\begin{itemize}

\item \textbf{Has the dataset already been used for tasks?}

Yes. It is used in this paper to evaluate query-driven multi-object tracking.

\item \textbf{Is there a repository listing papers using the dataset?}

No.

\item \textbf{What other tasks could this dataset support?}

The dataset can support research in vision-language reasoning, multimodal grounding, and agent-based visual decision-making.

\item \textbf{Could the dataset composition impact future uses?}

No known limitations are expected.

\item \textbf{Are there potential negative social impacts?}

No direct negative social impacts are anticipated.

\end{itemize}

\subsection{Distribution}

\begin{itemize}

\item \textbf{Will the dataset be distributed externally?}

Yes.

\item \textbf{How will the dataset be distributed?}

The dataset will be released through a public GitHub repository.

\item \textbf{Under what license will it be released?}

The dataset will be released under the Apache License.

\item \textbf{Are there any third-party restrictions?}

No.

\end{itemize}

\subsection{Maintenance}

\begin{itemize}

\item \textbf{Who will maintain the dataset?}

The authors of this paper.

\item \textbf{How can the maintainers be contacted?}

Contact details are provided in the paper.

\item \textbf{Will the dataset be updated?}

Yes. Updates and corrections will be made through the GitHub repository.

\item \textbf{Can others contribute to the dataset?}

Yes. Contributions and suggestions can be proposed through GitHub or by contacting the authors.

\end{itemize}


\begin{figure}[t]
\centering
\includegraphics[width=\linewidth]{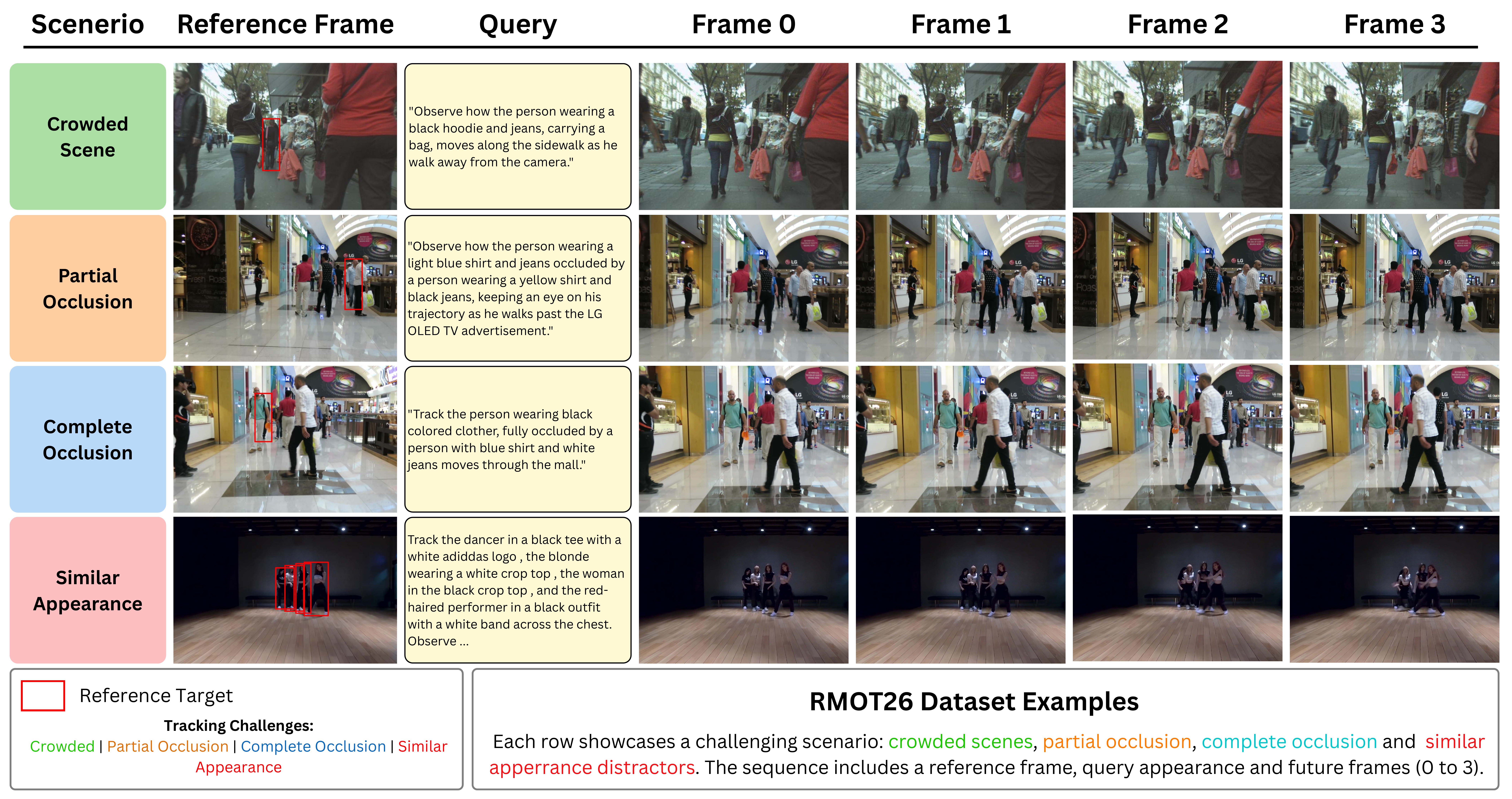}
\caption{Overview of the RMOT26 dataset components. The dataset contains four types of query scenarios: (1) crowded scenes, (2) partial occlusion, (3) complete occlusion, and (4) similar-appearance distractors. These cases require models to perform robust language-guided tracking under challenging visual conditions.}
\label{fig:dataset_components}
\end{figure}

\section{RMOT26 Dataset Statistics}
\label{appendix:dataset_statistics}

The \rmot dataset is constructed by curating video sequences from seven
existing multi-object tracking (MOT) benchmarks. To ensure fair evaluation
and avoid identity leakage between training and testing sets, all splits
are performed at the sequence level rather than the frame level.
\begin{table}[t]
\centering
\caption{Composition of the RMOT26 dataset derived from existing MOT benchmarks.}
\label{tab:dataset_composition}

\rowcolors{2}{gray!8}{white}

\begin{tabular}{lcccc}
\rowcolor{blue!20}
\textbf{Dataset} & \multicolumn{2}{c}{\textbf{Train}} & \multicolumn{2}{c}{\textbf{Test}} \\
\cmidrule(lr){2-3} \cmidrule(lr){4-5}
\rowcolor{blue!10}
 & \textbf{Sequences} & \textbf{Frames} & \textbf{Sequences} & \textbf{Frames} \\
\midrule

MOT17      & 253 & 1058 & 242 & 995 \\
MOT16      & 122 & 535 & 66  & 284 \\
MOT20      & 73 & 305 & 46  & 190 \\
DanceTrack & 962 & 3864 & 574 & 2275 \\
SportsMOT  & 73 & 297 & 2   & 8 \\
PETS       & 21 & 81 & 2   & 6 \\
DAMUNT     & 132 & 538 & 68  & 250 \\

\bottomrule
\end{tabular}

\end{table}
Figure~\ref{fig:dataset_components} illustrates the key components and
challenging scenarios represented in the dataset. In particular, the
dataset includes four types of query conditions: 
(1) crowded scenes, 
(2) partial occlusion, 
(3) complete occlusion, and 
(4) similar-appearance distractors. 
These scenarios require models to perform both spatial grounding and
temporal reasoning when tracking objects specified by natural language
queries.

\subsection{Dataset Composition}

Each instance in the dataset consists of a reference frame followed by a
short future sequence of 5--6 frames. The dataset is constructed using
sequences from multiple existing MOT benchmarks to ensure diversity in
scene types, motion patterns, and tracking difficulty.
Table~\ref{tab:dataset_composition} summarizes the dataset composition,
showing the contribution of each source dataset to the training and test
sets in terms of sequences and frames.

\subsection{Train / Test Splits}

The overall dataset statistics for the training and test splits are
presented in Table~\ref{tab:data_splits}. The training set contains
1,636 sequences and 6,678 frames, while the test set includes
1,000 sequences and 4,008 frames.
These splits ensure a balanced distribution of scenes and tracking
conditions while preventing sequence overlap between the training
and evaluation sets.

\begin{table}[h]
\centering
\caption{Overall RMOT26 dataset statistics for training and test splits.}
\label{tab:data_splits}

\rowcolors{2}{gray!8}{white}

\begin{tabular}{lccc}

\rowcolor{green!20}
\textbf{Split} & \textbf{Sequences} & \textbf{Frames} \\

\midrule

Train & 1636 & 6678 \\
Test  & 1000 & 4008 \\

\bottomrule
\end{tabular}

\end{table}

\section{Metrics and Implementation Details}
\label{appendix:implementation_details}

\begin{table}[t]
\centering
\caption{Evaluation metrics for RMOT26.}
\label{tab:metrics}
\small
\resizebox{\textwidth}{!}{
\begin{tabular}{l l}
\toprule
\rowcolor{headerblue}
\color{white}\textbf{Metric} & \color{white}\textbf{Description} \\
\midrule

\rowcolor{rowblue}
\textbf{IoU} & Spatial overlap between $B_t^{pred}$ and $B_t^{gt}$ per frame. \\

\rowcolor{rowgreen}
\textbf{Center Distance} & Euclidean distance between predicted and ground-truth centers $(x_t, y_t)$. \\

\rowcolor{rowyellow}
\textbf{Direction Consistency ($A_t$)} & Cosine similarity between predicted and ground-truth motion vectors $\Delta_t^{pred}$ and $\Delta_t^{gt}$. \\

\rowcolor{rowred}
\textbf{Speed Consistency ($S_t$)} & Gaussian penalty on motion magnitude difference $\|\Delta_t^{pred}\|$ and $\|\Delta_t^{gt}\|$. \\

\rowcolor{rowblue}
\textbf{MCP} & $\displaystyle \text{MCP} = \frac{1}{T-1} \sum_{t=2}^{T} A_t \cdot S_t$. \\

\bottomrule
\end{tabular}
}
\vspace{- 0.5 cm}
\end{table}
Table~\ref{tab:metrics} summarizes the evaluation metrics used throughout our experiments.
\noindent\textbf{Hardware setup.}
All model evaluations were conducted on a single NVIDIA H100 GPU (100GB), ensuring consistent hardware conditions across experiments. During evaluation, visual inputs and corresponding queries are processed by the selected \texttt{VLMM} agent, which produces structured outputs in a predefined \texttt{JSON} format.

\noindent\textbf{Training hyperparameters.}
Table~\ref{tab:hyperparam} summarizes the hyperparameters used to train \texttt{QTrack} on the \texttt{RMOT26} benchmark. Due to memory constraints, training was performed using only two frames per query. Training with the full sequence length may lead to different performance characteristics.

\paragraph{Train/test sequence length mismatch.}
Due to GPU memory constraints, \qtrack is trained using only two frames per query (the reference frame plus one future frame), while evaluation is performed on sequences of 5-6 frames. This mismatch means the model must generalize its learned motion reasoning to longer horizons than seen during training. Despite this constraint, \qtrack achieves the best performance across all metrics on \rmot, suggesting that the structured reward design and \tapo regularization provide generalizable motion-aware representations rather than overfitting to the two-frame training regime. Training with full sequence length remains an important direction for future work.

\begin{table}[tb]
\caption{Training hyperparameters for QTrack.}
\label{tab:hyperparam}
\centering

\resizebox{\textwidth}{!}{
\rowcolors{2}{lightgray}{white}

\begin{tabular}{l c|| l c}
\toprule
\rowcolor{headblue}
\textbf{Hyperparameter} & \textbf{Value} & \textbf{Hyperparameter} & \textbf{Value} \\
\midrule

\textit{max\_prompt\_length} & 1300 & \textit{max\_response\_length} & 2000 \\
\textit{max\_pixels} & 12845056 & \textit{use\_kl\_loss} & true \\
\textit{kl\_loss\_coef} & 5.0e-3 & \textit{fsdp.param\_offload} & false \\
\textit{fsdp.torch\_dtype} & null & \textit{offload.param\_offload} & true \\
\textit{offload.optimizer\_offload} & true & \textit{rollout.n} & 8 \\
\textit{limit\_images} & 2 & \textit{learning\_rate} & 1.0e-6 \\
\textit{weight\_decay} & 1.0e-2 & \textit{max\_grad\_norm} & 1.0 \\
\textit{global\_batch\_size} & 1 & \textit{adv\_estimator} & grpo \\
\textit{kl\_coef} & 0.0 & \textit{kl\_penalty} & kl \\
\textit{tapo\_strategy} & freeze & \textit{tapo\_keep\_prob} & 0.7 \\
\textit{tapo\_interval} & 2 & \textit{tapo\_kl\_weight} & 0.1 \\
\textit{rollout\_batch\_size} & 1 & \textit{temperature} & 1.0 \\
\textit{tensor\_parallel\_size} & 1 & \textit{total\_episodes} & 4 \\
\textit{n\_gpus\_per\_node} & 1 & & \\

\bottomrule
\end{tabular}
}

\end{table}

\subsection{Motion Consistency Precision (MCP)}
\label{appendix:mcp_metric}

Traditional tracking metrics primarily focus on spatial alignment (e.g., IoU) or identity preservation across frames. However, such metrics do not explicitly evaluate whether predicted trajectories follow physically consistent motion patterns over time. To address this limitation, we introduce \textbf{Motion Consistency Precision (MCP)}, a metric designed to measure the temporal coherence between predicted and ground-truth object trajectories.

MCP evaluates whether the predicted motion follows the same trajectory dynamics as the ground truth by jointly considering both the \emph{direction} and the \emph{speed} of motion. Intuitively, a reliable tracker should not only localize objects accurately in each frame but should also maintain motion patterns consistent with the underlying object dynamics.

\paragraph{Bounding-box convention.}
Ground-truth annotations are stored as
\[
\mathbf{b}^{xywh}
=
(x_{\mathrm{tl}},y_{\mathrm{tl}},w,h),
\]
where $(x_{\mathrm{tl}},y_{\mathrm{tl}})$ is the upper-left corner.
Model outputs use the corner-coordinate representation
\[
\widehat{\mathbf{b}}^{xyxy}
=
(\widehat{x}_1,\widehat{y}_1,\widehat{x}_2,\widehat{y}_2).
\]
Before computing rewards or evaluation metrics, a stored ground-truth
box is converted according to
\[
\phi(x_{\mathrm{tl}},y_{\mathrm{tl}},w,h)
=
(x_{\mathrm{tl}},y_{\mathrm{tl}},
 x_{\mathrm{tl}}+w,y_{\mathrm{tl}}+h).
\]
After conversion, all IoU, matching, center-distance, and motion
calculations operate on the common $xyxy$ representation. For a box
$\mathbf{b}=(x_1,y_1,x_2,y_2)$, its center is
\[
c(\mathbf{b})
=
\left(
\frac{x_1+x_2}{2},
\frac{y_1+y_2}{2}
\right).
\]

\paragraph{Motion Consistency Precision.}

The frame-level motion consistency score is defined as the product of direction and speed consistency:

\begin{equation}
\text{MCP}_t = A_t \cdot S_t.
\end{equation}

For a sequence of $T$ frames, the overall Motion Consistency Precision is computed as

\begin{equation}
\text{MCP} =
\frac{1}{T-1}
\sum_{t=2}^{T}
\text{MCP}_t.
\end{equation}

By construction, MCP lies within the range $[0,1]$, where higher values indicate stronger agreement between predicted and ground-truth motion trajectories. Unlike purely spatial metrics, MCP explicitly measures the \emph{temporal realism} of object motion, making it particularly suitable for evaluating tracking systems that must maintain coherent trajectories across time.

\paragraph{Relation to existing temporal metrics.}
Existing association-based metrics such as HOTA~\cite{luiten2021hota} decompose tracking performance into detection accuracy (DetA) and association accuracy (AssA). However, AssA measures identity consistency via overlap-based matching across frames and does not penalize physically implausible motion patterns. Similarly, TETA measures localization, association, and classification but does not enforce direction or speed coherence. MCP fills this gap by directly measuring whether predicted trajectories follow the same motion dynamics as ground-truth trajectories, making it uniquely suited to short-horizon query-conditioned tracking where identity switches are rare but spatial drift and motion inconsistency are the dominant failure modes.

\paragraph{Why not IDF1 or HOTA as primary metrics?}
Standard association metrics such as IDF1 and HOTA are designed for long-video, all-object tracking where identity switches and fragmentation are meaningful failure modes. In \rmot, each instance spans only 5-6 frames with explicit reference-frame grounding, meaning all models begin with a known target location. Under these conditions, identity switches are structurally rare across all evaluated models (exceeding 95\% IDF1 for most baselines), making association-based metrics uninformative for model discrimination. The dominant failure mode in our setting is spatial drift and motion inconsistency, which MCP directly quantifies. We therefore use MCP, MOTP, CLE, and NDE as primary metrics, and report HOTA and MOTA only in the traditional MOT comparison (Section~\ref{appendix:mot_eval_protocol}), where longer segments make association failures meaningful.


\section{Evaluation Protocol for Traditional MOT Comparison}
\label{appendix:mot_eval_protocol}
\begin{table}[t]
\centering
\caption{Comparison with specialized multi-object trackers and
reasoning-based trackers on MOT17 and DanceTrack. The results illustrate
the relative strengths of QTrack across localization, association, and
motion-consistency metrics.}
\label{tab:mot_17_dancetrack}
\begin{subtable}[t]{0.49\textwidth}
\centering
\caption{MOT17 Dataset}
\label{tab:mot17}
\resizebox{\linewidth}{!}{
\begin{tabular}{lcccc}
\toprule
\rowcolor{headerblue}
\color{white}\textbf{Model} &
\color{white}\textbf{MOTA} &
\color{white}\textbf{MOTP} &
\color{white}\textbf{HOTA} &
\color{white}\textbf{MCP} \\
\midrule
\rowcolor{rowblue}
MOTR & 0.61 & 0.81 & 0.22 & \underline{0.44} \\
BoostTrack++ & 0.63 & 0.76 & 0.38 & \textbf{0.44} \\
\rowcolor{rowgray}
TrackTrack & \textbf{0.75} & 0.50 & 0.23 & 0.29 \\
VisionReasoner & 0.64 & \underline{0.86} & \underline{0.60} & 0.21 \\
\midrule
\rowcolor{highlightgreen}
\textbf{QTrack} & \underline{0.69} & \textbf{0.87} & \textbf{0.69} & 0.26 \\
\bottomrule
\end{tabular}
}
\end{subtable}
\hfill
\begin{subtable}[t]{0.49\textwidth}
\centering
\caption{DanceTrack Dataset}
\label{tab:dancetrack}
\resizebox{\linewidth}{!}{
\begin{tabular}{lcccc}
\toprule
\rowcolor{headerblue}
\color{white}\textbf{Model} &
\color{white}\textbf{MOTA} &
\color{white}\textbf{MOTP} &
\color{white}\textbf{HOTA} &
\color{white}\textbf{MCP} \\
\midrule
\rowcolor{rowblue}
MOTR & 0.42 & 0.70 & 0.35 & 0.51 \\
MOTRv2 & 0.49 & 0.73 & 0.37 & \underline{0.52} \\
\rowcolor{rowgray}
TrackTrack & 0.36 & 0.73 & 0.40 & \textbf{0.55} \\
VisionReasoner & \underline{0.59} & \textbf{0.85} & \underline{0.61} & 0.26 \\
\midrule
\rowcolor{highlightgreen}
\textbf{QTrack} & \textbf{0.63} & \underline{0.83} & \textbf{0.66} & 0.35 \\
\bottomrule
\end{tabular}
}
\end{subtable}

\label{tab:mot_comparison}
\vspace{-0.3 cm}
\end{table}
To compare \qtrack with traditional multi-object tracking (MOT) methods, we design a query-based evaluation protocol derived from standard MOT benchmarks. Specifically, we use sequences from the MOT17~\cite{milan2016mot16} and DanceTrack~\cite{sun2022dancetrack} datasets, adapting them to reflect the query-driven nature of our task.
Traditional MOT benchmarks evaluate trackers that detect and track \emph{all} objects across an entire video. In contrast, \qtrack is designed to track only objects specified by a natural-language query. Direct evaluation on full MOT videos would therefore be inconsistent with the intended task. To address this mismatch, we construct \emph{query-specific tracking segments} from benchmark sequences so that both \qtrack and conventional trackers can be evaluated under comparable conditions.
For each sequence, we first select a reference frame containing the target object and generate a natural-language query describing that object. Starting from this reference frame, the tracker is required to follow only the queried target across subsequent frames. The predicted trajectories are then converted to the standard MOT annotation format, enabling the computation of common tracking metrics such as HOTA, MOTA, and MOTP.

This protocol allows \qtrack to be evaluated on widely used MOT datasets while preserving the query-based nature of the task. At the same time, it enables meaningful comparison with traditional trackers by isolating the trajectory of the queried object within a multi-object scene.
Notably, this evaluation setting introduces a stricter requirement for \qtrack. Instead of relying on global associations among all objects in the scene, the tracker must reason about the identity of the queried target and maintain its trajectory over time. Despite this constraint, \qtrack achieves competitive or superior performance on standard MOT metrics, demonstrating that query-aware reasoning can improve robustness in complex tracking scenarios.
Overall, this protocol bridges traditional MOT evaluation with query-based tracking, enabling fair and practical comparisons between the two paradigms.

Table~\ref{tab:mot_comparison} presents the quantitative comparison between \qtrack and traditional MOT methods on query-specific segments from MOT17 and DanceTrack. On MOT17, \qtrack achieves the best MOTP  (0.87) and HOTA (0.69) while maintaining competitive MOTA (0.69). On  DanceTrack, \qtrack achieves the best MOTA (0.63) and HOTA (0.66). These results confirm that query-conditioned reasoning does not sacrifice trajectory quality relative to methods that track all objects indiscriminately.


\section{Comparison with Fine-Tuned Models}
\label{appendix:finetunned_models}
\begin{table*}[t]
\centering
\caption{Comparison of VLLM models after fine-tuning.}
\label{tab:fine_tunned}

\small

\rowcolors{2}{lightgray}{white}

\begin{tabular}{lccccc}
\toprule
\rowcolor{headblue}
\textbf{Model} & \textbf{Params} & \textbf{MCP$\uparrow$} & \textbf{MOTP$\uparrow$} & \textbf{MOTA$\uparrow$} & \textbf{NDE$\downarrow$} \\
\midrule

Qwen 2.5 VL Instruct~\cite{qwen2.5-VL} & 3B  & 0.14 & \textbf{0.76} & -0.51 & 3.41 \\
Gemma3~\cite{team2025gemma} & 4B & 0.18 & 0.73 & -0.16 & 0.95 \\
Visionreasoner~\cite{liu2025visionreasoner} & 3B & \underline{0.22} & 0.65 & \underline{0.01} & \underline{0.76} \\

\midrule
\rowcolor{lightgreen}
\textbf{QTrack (Ours)} & \textbf{3B} & \textbf{0.30} & \underline{0.75} & \textbf{0.21} & \textbf{0.39} \\

\bottomrule
\end{tabular}

\end{table*}
Table~\ref{tab:fine_tunned} evaluates fine-tuned VLLMs under identical training settings. QTrack achieves the highest MCP (0.30), highest MOTA (0.21), and lowest NDE (0.39), highlighting the importance of reward design and temporal regularization beyond supervised fine-tuning.


\begin{figure}[!htbp]
\centering
\includegraphics[width=\linewidth]{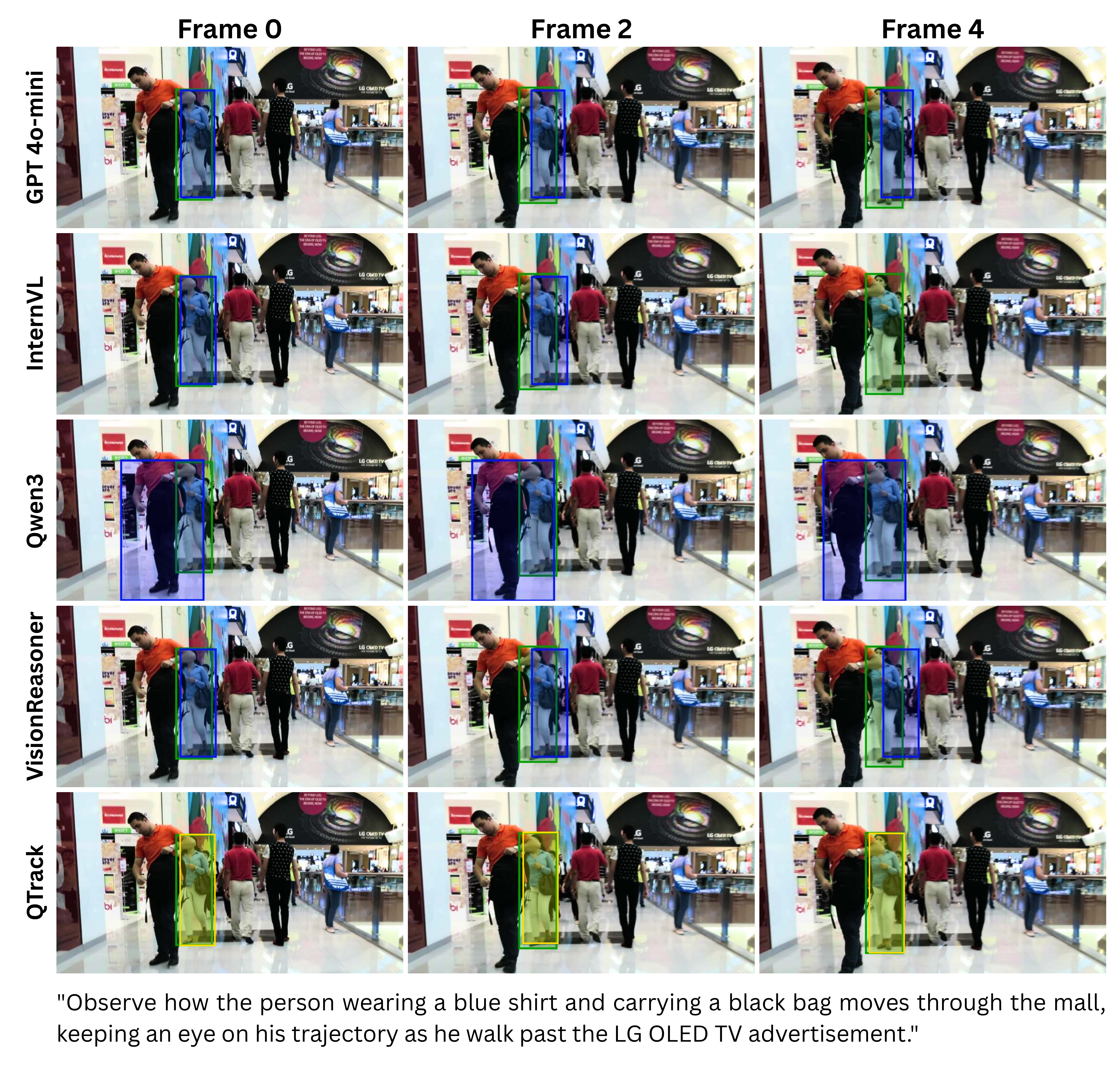}
\caption{Qualitative comparison for single-object tracking across multiple frames. \qtrack produces more accurate and stable bounding boxes over time, while baseline VLLM models often exhibit drifting or inconsistent localization as the sequence progresses.}
\label{fig:ablation_1}
\end{figure}

\section{Additional Qualitative Results}
\label{appendix:qualatative}

\begin{figure}[!htbp]
\centering
\includegraphics[width=\linewidth]{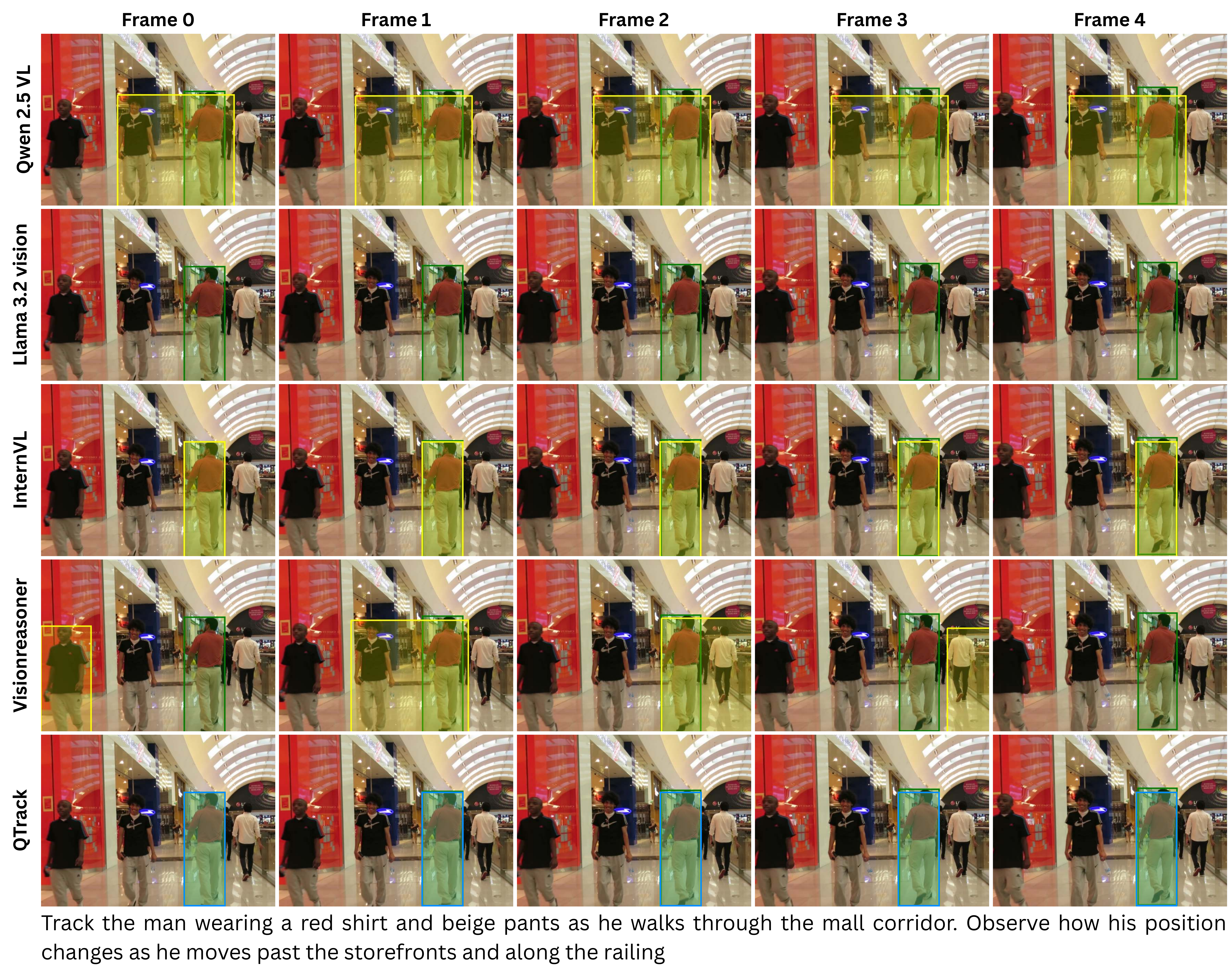}
\caption{Additional qualitative comparison of different VLLM models on challenging tracking scenarios. Our method demonstrates stronger temporal consistency and better object localization compared to baseline models.}
\label{fig:ablation_2}
\end{figure}

We present additional qualitative comparisons to further analyze the behavior of \qtrack under different tracking scenarios. Figures~\ref{fig:ablation_1},~\ref{fig:multi_object_failure}, and~\ref{fig:ablation_2} compare the predictions of several VLLM-based baselines with our proposed method across multiple frames. As illustrated, \qtrack produces more stable and accurate trajectories across time, particularly in challenging situations involving motion, occlusion, and appearance similarity. In contrast, baseline models often exhibit drifting predictions or inconsistent localization across frames.
Figure~\ref{fig:multi_object_failure} presents a qualitative comparison for multi-object tracking. VisionReasoner fails to detect and track all objects across frames, whereas \qtrack successfully localizes and tracks all queried objects in every frame.

\begin{figure}[!htbp]
\centering
\includegraphics[width=\linewidth]{Images/multi_object_failure_case.pdf}
\caption{ In this figure, Visionreasoner is not able to detect and track all the objects in the frames, but
QTrack can detect and track all objects in all frames}
\label{fig:multi_object_failure}
\end{figure}

\begin{figure}[t]
\centering
\includegraphics[width=\linewidth]{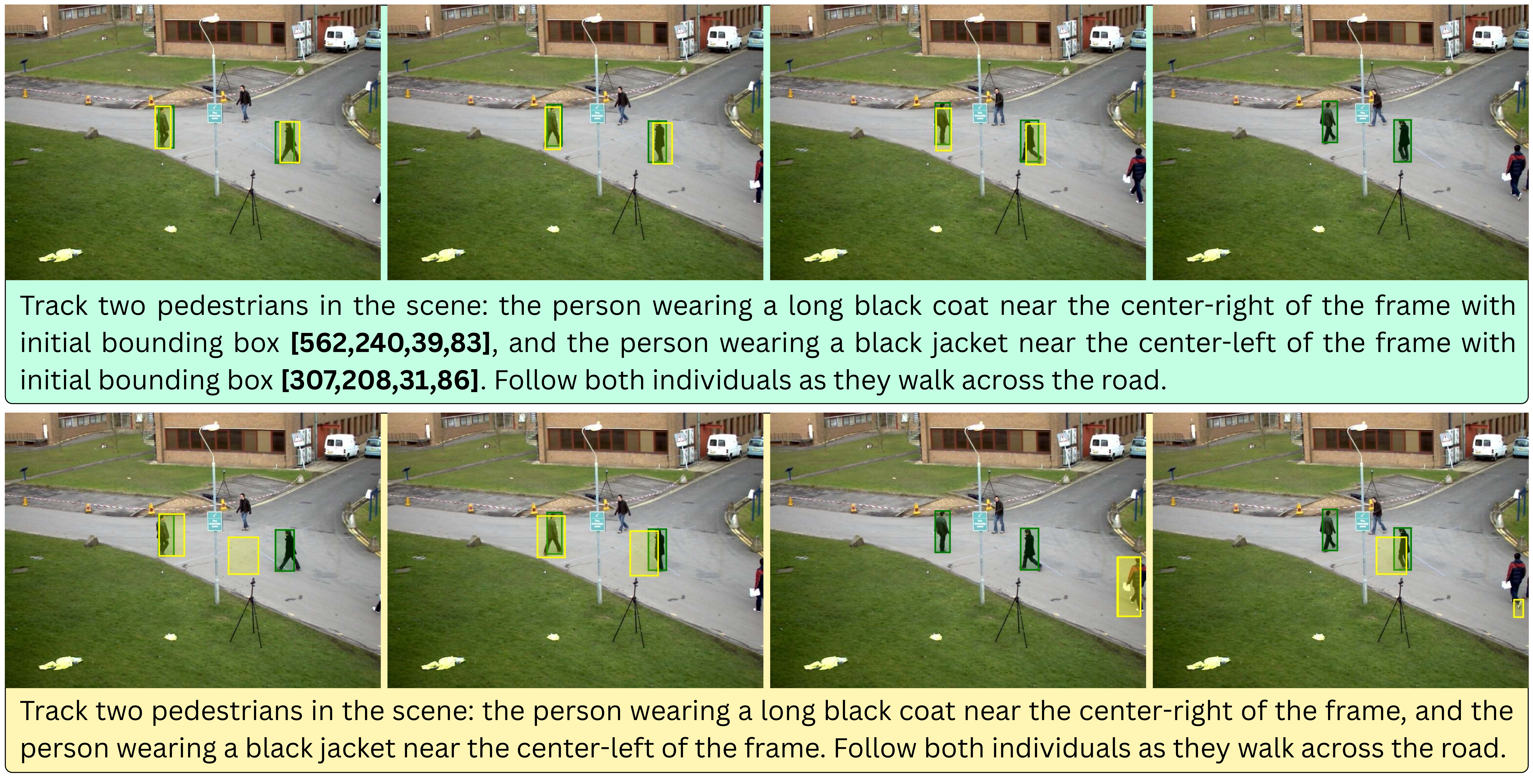}
\caption{Effect of spatial initialization. Tracking is reliable with an initial bounding box, but localization becomes difficult when relying solely on language queries.}
\label{fig:with_without_prompt}
\vspace{-0.5 cm}
\end{figure}

Figure~\ref{fig:with_without_prompt} investigates the impact of providing an initial bounding box in the reference frame. We evaluate two settings: (1) the model receives both a natural-language query and the initial bounding box of the target object, and (2) the model receives only the natural-language query without any spatial initialization. When the initial bounding box is provided, the model can reliably maintain the trajectory of the queried object across subsequent frames. However, when the model relies solely on the textual query, it occasionally struggles to precisely localize the target object in the first frame. This highlights an important limitation of current vision-language models: while they can reason about object descriptions, precise spatial grounding from language alone remains challenging in complex scenes.

\begin{figure}[!htbp]
\centering
\includegraphics[width=\linewidth]{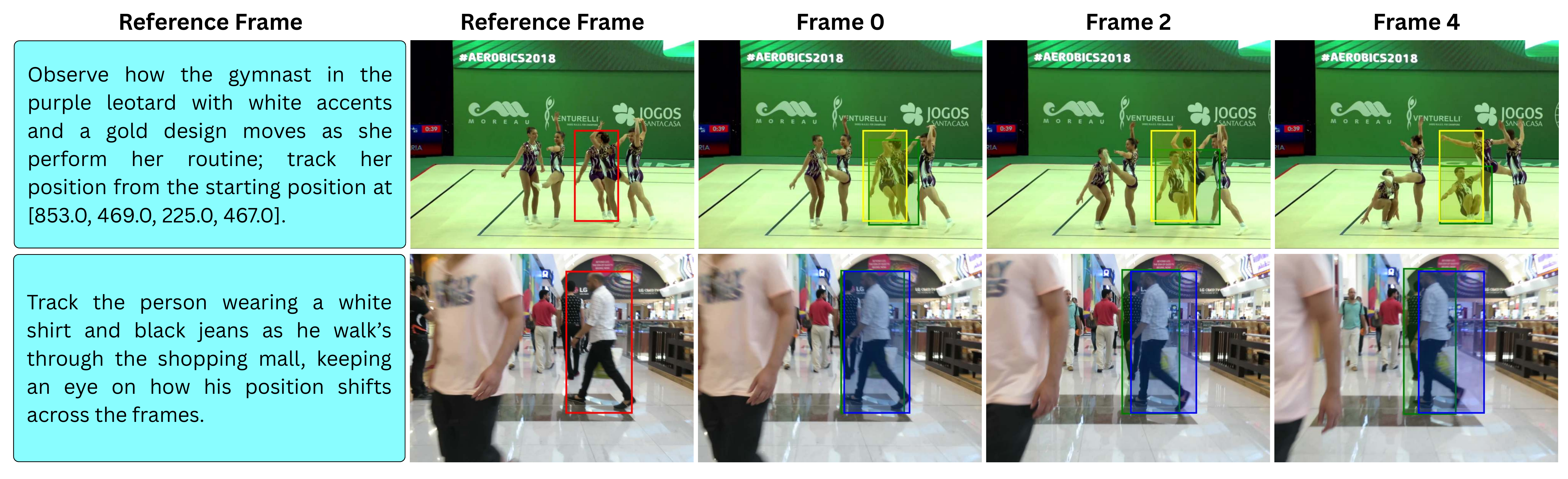}
\caption{This figure shows failure cases of QTrack, we can see that for some sequences QTrack is unable to predict next bounding box for frames, it repeatedly predicts the same initial bounding box for all future frames.}
\label{fig:fail_cases}
\vspace{-0.4 cm}
\end{figure}
\section{Failure Cases}
\label{appendix:fail_cases}
Figure~\ref{fig:fail_cases}, shows failure cases of QTrack; we can see that for some sequences, QTrack is unable to predict the next bounding box for frames; it repeatedly predicts the same initial bounding box for all future frames.


\section{Prompt Templates for Baseline Models}
\label{appendix:prompt}

To ensure a fair comparison, all baseline models are evaluated using the same input information and prompt structure.

\paragraph{Input format.}
Each model receives the following inputs:
\begin{itemize}
    \item Video frames from the sequence
    \item A natural-language query describing the target object
    \item A reference bounding box indicating the initial location of the object in the reference frame
\end{itemize}

\paragraph{Prompt template.}
We use a unified prompt template for all baseline models. The template instructs the model to track the queried object across the provided video frames and output the predicted bounding boxes in a structured JSON format.

\begin{tcolorbox}[colback=gray!5,colframe=black,title=Prompt Template]
\small
\begin{verbatim}
{textual_query}

Initially, object {obj_id} is located at {initial_bbox}
in the reference image.

The following images correspond to consecutive
frames of a video.

Please output the bounding box for the object in
each frame using the following format:

[{"frame": N, "bbox": [x1, y1, x2, y2]}, ...]

Do not include any additional text.
\end{verbatim}
\end{tcolorbox}

This prompt ensures that all models receive identical instructions and output constraints. By enforcing a structured JSON response format, we simplify downstream parsing and ensure consistent evaluation across different baseline models.

\section{Reasoning Quality Evaluation Details}
\label{appendix:reasoning}

\paragraph{Why Reasoning?}

To check if reasoning is adding any contribution to results we added a direct trajectory-only ablation in Table~\ref{tab:why_reasoning}. Removing structured reasoning degrades every metric: MCP decreases from 0.30 to 0.28, MOTP decreases from 0.75 to 0.72, CLE increases from 44.61 to 129.5, approximately a 2.9× increase, NDE increases from 0.39 to 0.81, approximately a 2.1× increase.

\begin{table}[h]
  \centering
  \caption{Ablation for reasoning}
  \label{tab:why_reasoning}
  \setlength{\tabcolsep}{5pt}
  \begin{tabular}{lcccc}
    \toprule
    \rowcolor{headblue}
    \textbf{Model} & \textbf{MCP$\uparrow$} & \textbf{MOTP$\uparrow$} & \textbf{CLE$\downarrow$} & \textbf{NDE$\downarrow$} \\
    \midrule
    
    QTrack with structured reasoning & \textbf{0.30}	 & \textbf{0.75} & \textbf{44.61} & \textbf{0.39} \\
    QTrack without structured reasoning & 0.28 & 0.72 & 129.5 & 0.81 \\
    
    \bottomrule
  \end{tabular}
\end{table}

The structured reasoning field gives the policy an intermediate representation for binding: language-specified attributes, selected target identity, motion direction, visibility changes, object interactions, final trajectory coordinates.
The ablation suggests that this structure is functionally useful rather than purely post-hoc. We do not claim that every generated rationale is perfectly faithful. This is why the paper separately analyzes 200 reasoning traces and reports unsupported-motion and weak-causal-justification failures. The two analyses answer different questions: (1) the correlation analysis measures whether better individual reasoning traces are associated with better predictions (2) the trajectory-only intervention tests whether the structured reasoning-output formulation contributes to tracking performance.

\paragraph{Scoring protocol.}
We evaluate $200$ test instances sampled with stratification across four IoU tiers: easy (${\geq}0.7$, $n{=}93$), medium ($[0.4,0.7)$, $n{=}56$), hard ($[0.1,0.4)$, $n{=}28$), and failure ($<0.1$, $n{=}23$). Each reasoning trace is scored on four dimensions (0--2 each): (A)~\emph{Object identification} explicit reference to target identity or initial location; (B)~\emph{Temporal reasoning} mention of motion direction, continuity, or occlusion; (C)~\emph{Causal justification} explanation of \emph{why} the target matches the query beyond appearance; (D)~\emph{Faithfulness} geometric consistency between stated motion direction and predicted trajectory. RQS is defined as:
\begin{equation}
  \text{RQS} = \frac{1}{8N}\sum_{i=1}^{N}
    \bigl(r_i^{\text{rel}}+r_i^{\text{temp}}+
          r_i^{\text{causal}}+r_i^{\text{faith}}\bigr).
\end{equation}

\begin{table}[h]
  \centering
  \caption{RQS and faithfulness by IoU tier ($n{=}200$).}
  \label{tab:reasoning}
  \setlength{\tabcolsep}{5pt}
  \begin{tabular}{lcccc}
    \toprule
    Tier & $n$ & IoU & RQS & Faithfulness \\
    \midrule
    Easy   (${\geq}0.7$)  & 93 & 0.882 & 0.593 & 0.661 \\
    Medium ($[0.4,0.7)$)  & 56 & 0.599 & 0.554 & 0.580 \\
    Hard   ($[0.1,0.4)$)  & 28 & 0.278 & 0.585 & 0.643 \\
    Failure ($<0.1$)      & 23 & 0.021 & 0.489 & 0.457 \\
    \midrule
    All & 200 & 0.619 & 0.569 & 0.280$^{\dagger}$ \\
    \bottomrule
  \end{tabular}\\[4pt]
  \small $^{\dagger}$Fraction with faithfulness $\geq 1.5$.
  RQS\,$\leftrightarrow$\,IoU: $r{=}{+}0.158$, $p{=}0.025$.
  RQS\,$\leftrightarrow$\,MCP: $r{=}{-}0.10$, $p{=}0.16$ (n.s.).
\end{table}

\paragraph{Failure mode distribution.}
Among $51$ low-IoU samples (hard + failure tiers): weak causal justification 25 (49\%), directional hallucination 9 (18\%), motion failure 9 (18\%), appearance bias 8 (16\%).

\end{document}